\documentclass[acmsmall]{acmart}

\usepackage{amsmath,amsfonts}
\usepackage{algorithmic}
\usepackage{algorithm}
\usepackage{array}
\usepackage{textcomp}
\usepackage{subfigure}
\usepackage{stfloats}
\usepackage{url}
\usepackage{verbatim}
\usepackage{graphicx}
\usepackage{color}
\usepackage{cases} 
\usepackage{multirow}
\usepackage{soul}

\AtBeginDocument{%
  \providecommand\BibTeX{{%
    \normalfont B\kern-0.5em{\scshape i\kern-0.25em b}\kern-0.8em\TeX}}}

\setcopyright{acmcopyright}
\copyrightyear{2018}
\acmYear{2018}
\acmDOI{XXXXXXX.XXXXXXX}

\acmJournal{JACM}
\acmVolume{37}
\acmNumber{4}
\acmArticle{111}
\acmMonth{8}

\newcommand{\sysname}{{TKN}}
\newcommand{\etal}{{\it et al.}}

\begin{document}

%%
%% The "title" command has an optional parameter,
%% allowing the author to define a "short title" to be used in page headers.
\title{TKN: Transformer-based Keypoint Prediction Network For Real-time  Video Prediction}

%%
%% The "author" command and its associated commands are used to define
%% the authors and their affiliations.
%% Of note is the shared affiliation of the first two authors, and the
%% "authornote" and "authornotemark" commands
%% used to denote shared contribution to the research.
\author{Haoran Li}
% \authornote{Both authors contributed equally to this research.}
\email{lhr123@mail.ustc.edu.cn.com}
% \orcid{1234-5678-9012}
\affiliation{%
  \institution{University of Science and Technology of China}
  % \streetaddress{P.O. Box 1212}
  \city{Hefei}
  % \state{Anhui}
  \country{China}
  % \postcode{43017-6221}
}
\author{Xiaolu Li}
% \authornotemark[1]
\email{xiaoluli0718@mail.ustc.edu.cn}
\affiliation{%
  \institution{University of Science and Technology of China}
  % \streetaddress{P.O. Box 1212}
  \city{Hefei}
  % \state{Anhui}
  \country{China}
  % \postcode{43017-6221}
}
\author{Yihang Lin}
% \authornotemark[1]
\email{lyh1998@mail.ustc.edu.cn}
\affiliation{%
  \institution{University of Science and Technology of China}
  % \streetaddress{P.O. Box 1212}
  \city{Hefei}
  % \state{Anhui}
  \country{China}
  % \postcode{43017-6221}
}
\author{Yanbin Hao}
% \authornotemark[1]
\email{haoyanbin@hotmail.com}
\affiliation{%
  \institution{University of Science and Technology of China}
  % \streetaddress{P.O. Box 1212}
  \city{Hefei}
  % \state{Anhui}
  \country{China}
  % \postcode{43017-6221}
}
\author{Haiyong Xie}
% \authornotemark[1]
\email{haiyong.xie@ieee.org}
\affiliation{%
  \institution{Adv. Innovation Center for Human Brain Protection, Capital Medical University}
  % \streetaddress{P.O. Box 1212}
  \city{Beijing}
  % \state{Anhui}
  \country{China}
  % \postcode{43017-6221}
}

\author{Pengyuan Zhou}
\email{pengyuan.zhou@ece.au.dk}
%\authornotemark[1]
\affiliation{%
  \institution{Aarhus University}
  % \streetaddress{P.O. Box 1212}
  \city{Aarhus}
  % \state{Anhui}
  \country{Denmark}
  % \postcode{43017-6221}
}

\author{Yong Liao}
\authornotemark[1]
\email{yliao@ustc.edu.cn}
\affiliation{%
  \institution{University of Science and Technology of China}
  % \streetaddress{P.O. Box 1212}
  \city{Hefei}
  % \state{Anhui}
  \country{China}
  % \postcode{43017-6221}
}

%%
%% By default, the full list of authors will be used in the page
%% headers. Often, this list is too long, and will overlap
%% other information printed in the page headers. This command allows
%% the author to define a more concise list
%% of authors' names for this purpose.
\renewcommand{\shortauthors}{Trovato and Tobin, et al.}

%%
%% The abstract is a short summary of the work to be presented in the
%% article.

%%%%%%%%% ABSTRACT
\begin{abstract}
Video prediction is a complex time-series forecasting task with great potential in many use cases. However, conventional methods overemphasize accuracy while ignoring the slow prediction speed caused by complicated model structures that learn too much redundant information with excessive GPU memory consumption. Furthermore, conventional methods mostly predict frames sequentially (frame-by-frame) and thus are hard to accelerate. Consequently, valuable use cases such as real-time danger prediction and warning cannot achieve fast enough inference speed to be applicable in reality.  %Conventional methods have been ignoring the slow prediction speed and excessive GPU consumption caused by complex structure employment and redundant information learning in the pursuit of high precision results. 
%In this paper, 
Therefore, we propose a transformer-based keypoint prediction neural network (\sysname), an unsupervised learning method that boost the prediction process via constrained information extraction and parallel prediction scheme. %\textcolor{red}{Comment 1 from Reviewer 2}. 
TKN is the first real-time video prediction solution to our best knowledge, while significantly reducing computation costs and maintaining other performance.   %nearly the same accuracy and quality of prediction as the state-of-the-art approaches.
%To our best knowledge, \sysname\ is the first real-time prediction solution. %\textcolor{blue}{\sysname\ outperforms state-of-the-art approaches in terms of better keypoint capturing and image reconstruction in video prediction field.}  \textcolor{red}{Comment 5 from Reviewer 2} 
Extensive experiments on KTH and Human3.6 datasets demonstrate that \sysname\ predicts 11 times faster than existing methods while reducing memory consumption by 17.4\% and achieving state-of-the-art prediction performance on average. 
 %decreasing the processing delay by up to 295\%  
\end{abstract}

%%
%% The code below is generated by the tool at http://dl.acm.org/ccs.cfm.
%% Please copy and paste the code instead of the example below.
%%


%%
%% Keywords. The author(s) should pick words that accurately describe
%% the work being presented. Separate the keywords with commas.

\received{20 February 2007}
\received[revised]{12 March 2009}
\received[accepted]{5 June 2009}

\maketitle
%%
%% This command processes the author and affiliation and title
%% information and builds the first part of the formatted document.

%预测未来一直是人类梦寐以求的能力，它可以让我们抓住机会对未来要发生的事情做好准备。在一些突发情况中，不仅需要正确的预测，更需要快速实时的去预测，比如说在一辆疾驰的汽车中，司机的应对危险的时间通常在2.3-3秒，如果我们预测系统预测时间太长就会造成严重的危险。现在随着人工智能在计算机视觉领域的的发展，预测未来正逐渐成为现实，视频预测这个领域正是可以把未来的以视频的方法展现出来，但是目前的方法都专注于预测的精准度，忽略了关键的预测速度，所以我们通过高效的关键点提取，使用可以并行计算的自注意力模块实现了实时的视频预测。
%视频预测是根据已知的视频帧序列，预测后续的视频帧序列，它隶属于时间序列预测问题，最早应用于降水的雷达回波图的预测，后面逐渐应用到了人类活动中。目前的主流视频预测方法可以分为两种：第一类，是改进著名的递归神经网RNN，使之它可以更好的捕捉编码相邻帧之间的规律。然而基于RNN的方法将隐藏层信息在序列间按顺序传递，这就会导致最初的信息在传递过程中逐渐消失，从而导致所谓的短时记忆，影响长序列预测的精度。第二类是先对视频帧进行比较分析，一般会把视频帧解纠缠分成运动部分和静止部分，然后对两个模块分别进行预测处理。
%上面所说的两类视频预测方法为了获取精确的预测结果，大都会对每一帧提取复杂的特征，然后对这些特征送入预测网络进行预测，一般单张帧的特征就会达到上万字节，这样无论是在提取特征的模块还是在预测模块都会造成大量的浮点计算，所以无论是在训练还是测试中都会有大量的时间和显存上的消耗。基于以上挑战，我们希望提出一种能够极大减少计算量从而减少视频预测时间以及所消耗显存的方法。

%%%%%%%%% BODY TEXT
\section{Introduction}
\label{sec:intro}
Predicting the future has always been a coveted ability that allows users to be well prepared for upcoming events. With the advancement of artificial intelligence in the field of computer vision, the ability to predict the future is gradually becoming a reality. One of the most popular methods is video prediction, which predicts subsequent video frame sequences based on prior ones. It belongs to the time series prediction problem and was initially applied to the prediction of radar echo maps of precipitation \cite{shi2015convolutional} then to human activities \cite{wang2017predrnn,wang2018eidetic}. 
Current mainstream video prediction methods can be divided into two categories. %\textcolor{red}{according to their structure}. 
The first is to improve the well-known recurrent neural network (RNN) to accurately capture the inter-frame pattern \cite{wang2018eidetic,wang2021predrnn,wang2017predrnn}. However, RNN-based methods often gradually lose the initial information during the sequential information transmission across the hidden layers, resulting in the so-called \emph{short memory} that negatively impacts the long sequence prediction accuracy \cite{zhao2020rnn}. The second category divides a video frame into a moving portion and a stationary portion and then predicts the two portions separately \cite{ying2018better,guen2020disentangling,minderer2019unsupervised,gao2021accurate}.  
%shows the future in a video method. 
\par Most existing works focus on improving accuracy by a few percentage points while ignoring the prediction speed, which is actually crucial for many real-time applications. For instance, in a speeding car, the driver can typically afford a reaction time to danger below 3 seconds~\cite{mcgehee2000driver} otherwise would face a grave risk. Assume we want to predict the video frames for the next 3 seconds with a typical vehicular front camera rate of 60 frames per second (fps), the video prediction method has to reach at least 180\~fps to finish the prediction within one second. However, existing methods can normally support a frame rate only up to 80 to 100 fps~\cite{guen2020disentangling,akan2021slamp,gao2021accurate}, which can barely help in reality. The reason is threefold: 1) existing methods extract complex features for the sake of higher accuracy, resulting in an excessive number of floating point operations\cite{wang2018eidetic,akan2021slamp,chen2020long}; 2) they waste considerable time on learning similar background information often shared by consecutive frames~\cite{schuldt2004recognizing,h36m_pami}; 3) they use a sequential prediction process where the next frame's input depends on the previous frame's output. Consequently, these methods are poor at processing efficiency and can not predict multiple frames in parallel.
As such, we propose a Transformer-based Keypoint extraction neural Network (TKN), which is an unsupervised learning method consisting of a keypoint detector and predictor. \sysname\ can predict video frames by predicting only the keypoints. The keypoint detector extracts feature data for only a few tens of bytes and achieves temporal parallelism, hence greatly reducing the number of floating-point operations, the prediction time, and memory consumption. The predictor further accelerates the process by gathering global attention information in a parallel manner via a self-attention mechanism without disregarding past information. Our contributions are threefold as follows.

%model 改完以后再改
\begin{itemize}
    \item \sysname\ incorporates the advantages of both Keypoint and Transformer structures to guarantee high prediction accuracy, fast training and testing, and low memory consumption. In order to accurately predict videos that contain frequent changes, we additionally propose a sequential variation of \sysname\ called \sysname-Sequential.
    \item The keypoint detector of \sysname\ predict multiple frames in parallel and outperforms keypoint-based state-of-the-art (SOTA) methods in the field of video prediction in terms of keypoint capture and frame reconstruction, resulting in increasing SSIM by 6.3\% and PSNR by 7.5\% with 88.1\% fewer floating-point operations.%%\textcolor{red}{R2C5}
    \item Extensive experimental evaluations have demonstrated the superiority of \sysname\, which achieves a prediction speed of 1176~fps and thus realizing the first real-time video prediction to our best knowledge. % and  \pz{add some niubi result numbers here, especially those more than 100\%} \textcolor{red}{
    Compared to existing methods, TKN is 11 times faster at prediction while reducing 17.4\% GPU memory consumption. As such, \sysname\ lays the groundwork for future real-time multimedia technologies.
\end{itemize}

\section{Related Works}
\label{sec:related}

Unsupervised methods can reduce the cost of manual annotation which is a common requirement for video datasets.

\noindent\textbf{Unsupervised keypoint learning.}
Due to the similarity of pixels in consecutive video frames, the keypoints in each frame can be learned via unsupervised reconstruction of the other frames. Jakab~\etal~\cite{jakab2018conditional} propose to learn the object landmarks via conditional image generation and representation space shaping.
%
%Minderer~\etal~\cite{minderer2019unsupervised} adopted stochastic dynamics learning and applied keypoints to video prediction for the first time. It greatly reduces the computational complexity and enables real-time prediction. 
Minderer~\etal~\cite{minderer2019unsupervised} introduce keypoints to video prediction using stochastic dynamics learning for the first time, which drastically reduces computational complexity.
Gao~\etal~\cite{gao2021accurate} applied grids on top of~\cite{minderer2019unsupervised} for a clearer expression of the keypoint distribution.

\begin{figure} [t!]
\centering
\subfigure[]{
\begin{minipage}[t]{0.45\linewidth}
\centering
\includegraphics[width=\textwidth]{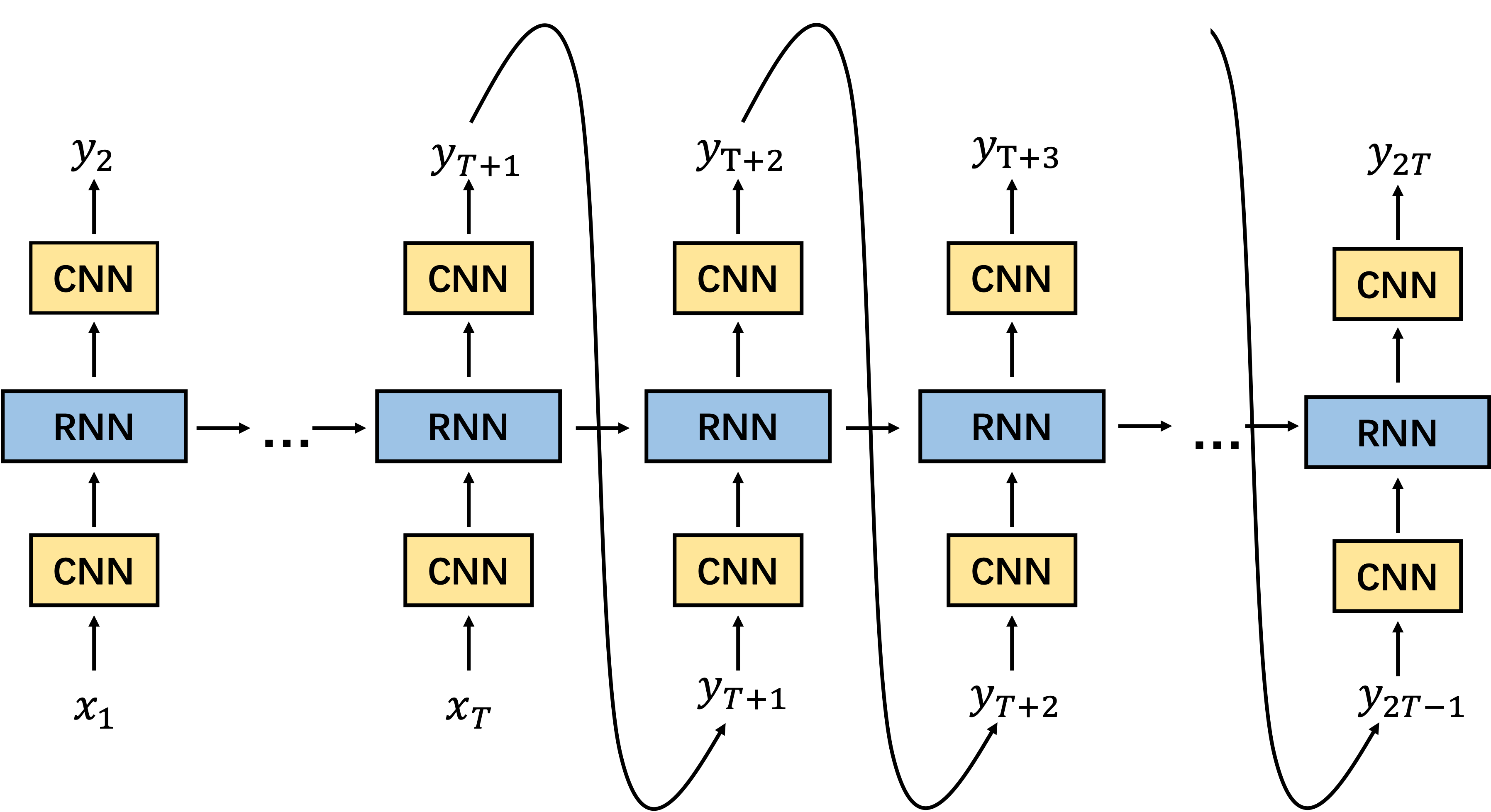}
%\caption{fig1}
\label{fig:otherstrutures}
\end{minipage}%
}
\subfigure[]{
\begin{minipage}[t]{0.45\linewidth}
\centering
\includegraphics[width=\textwidth]{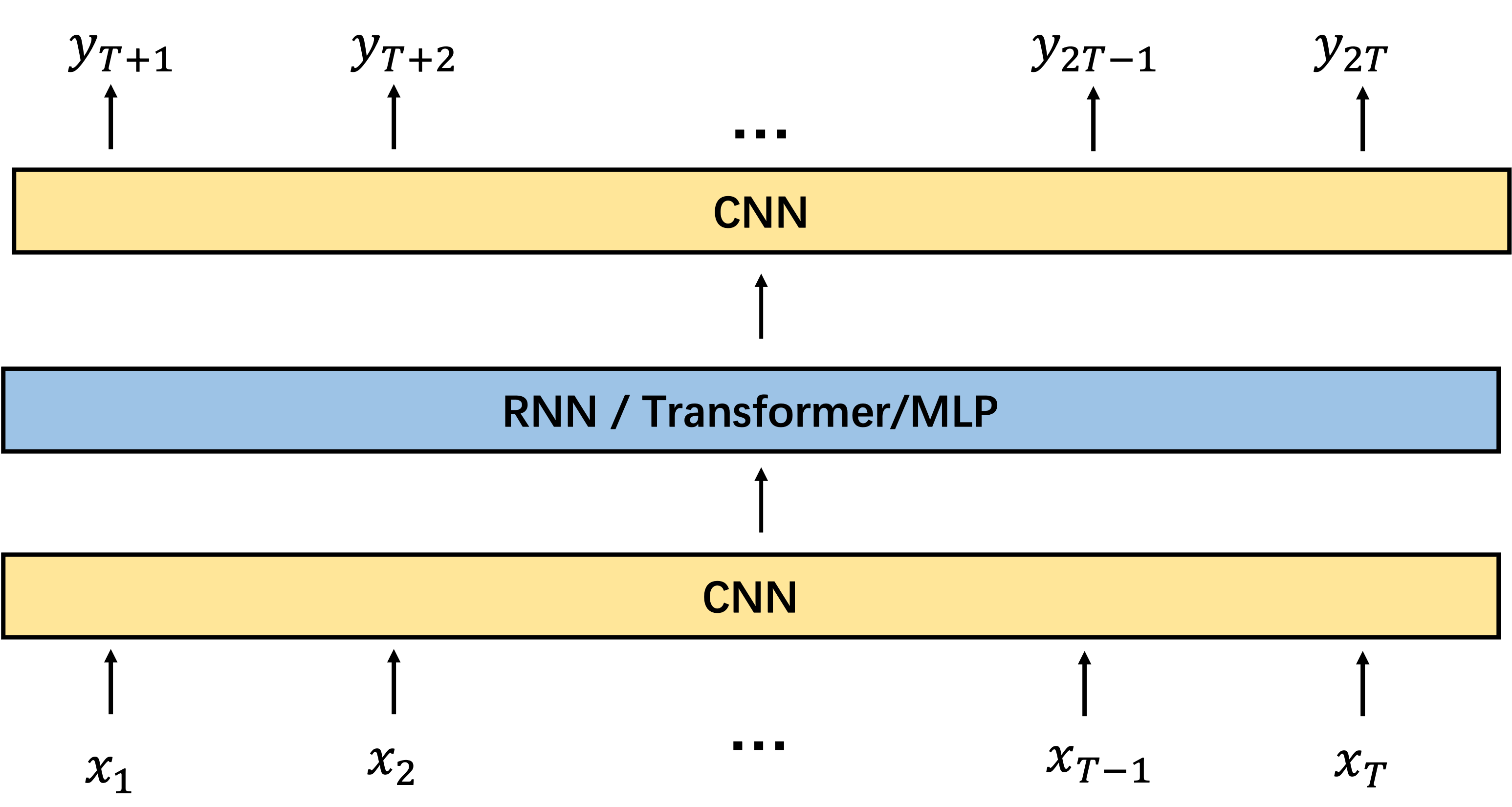}
%\caption{fig1}
\label{fig:mystructure}
\end{minipage}%
}

	\caption{(a) The sequential prediction scheme generally takes a long time to predict frames due to the sequential scheme. (b) The parallel prediction scheme we propose can greatly accelerate the prediction speed.}%represents the majority of unsupervised video prediction processes, which can only be performed frame by frame and can not  predict  long periods of  a short time. (b) is a multi-frame prediction process, which saves large amount of time in the feature extraction and output stages compared to (a).}
	\label{fig3} 
\end{figure}

\noindent\textbf{Unsupervised video prediction}
uses the pixel values of the video frames as the labels for unsupervised prediction. Existing studies can be classified into two categories, as shown by Fig.~\ref{fig:otherstrutures}. 
The first category of works focuses on improving the performance of the well-known RNN by adapting the intermediate recurrent structure~\cite{wang2018eidetic,wang2021predrnn,oliu2018folded,wang2018predrnn++,castrejon2019improved}. For example, E3D-LSTM~\cite{wang2018eidetic} integrates 3DCNN with LSTM to extract short-term dependent representations and motion features. PredRNN~\cite{wang2021predrnn} enables the cross-level communication for the learned visual dynamics by propagating the memory flow in both bottom-up and top-down orientations. 
The second category focuses on disentangling the dynamic objects and the static background in the video frames, mostly by adapting the CNN structure~\cite{ying2018better,guen2020disentangling,denton2017unsupervised,blattmann2021understanding,xu2019unsupervised}.
For instance, DGGAN~\cite{ying2018better} trains a multi-stage generative network for prediction guided by synthetic inter-frame difference. PhyDNet~\cite{guen2020disentangling} uses a latent space to untangle physical dynamics from residual information. %Our work can also be classified in the second category.

The methods in both categories use so-called ``sequential prediction'', that is, using the previous prediction frame as the input frame for the next round of prediction. The prediction speed is proportional to the number of frames to be predicted and thus leads to an intolerably long delay for long-term prediction. 
Therefore, we propose a parallel prediction scheme, as shown in Fig.~\ref{fig:mystructure}, to extract the features of multiple frames and output multiple predicted frames in parallel, which greatly accelerates the prediction process.
%\par \textcolor{red}{The prediction process of both categories of methods can mostly be represented by the \ref{fig:otherstrutures}. For the first category of methods , they focus on improving the intermediate recurrent structure, while the second category of methods mostly change the CNN structure to better disentangle the static and dynamic information in videos. They both use the prediction frame as the next input frame for the next round of prediction and this prediction process can be called frame-by-frame prediction. Obviously, this prediction inference time is T times that of a single prediction process time, and as mentioned before, if the network structure is particularly complex, the time of a single prediction process is very long, so it is impossible to predict long video sequences in a very short time. Therefore, we propose a multi-frame prediction process, as shown in \ref{fig:mystructure} , which can extract the features of  video frames at the same time, and output the predicted T frames at the same time, which can greatly reduce the prediction time.}
%目前的方法按照结构可以分为两种，这两类方法的预测流程大都可以用图1中a的来表示，对于第一类方法是改进中间层的预测结构，而第二类方法多是改变输入和输出的CNN结构。他们流程都是使用预测帧作为下一输入帧从而进行下一轮的预测，这种预测流程我们可以称之为逐帧预测，很明显，这种预测推理时间是单次预测的T倍，如之前所说，如果网络结构特别复杂，导致单次预测流程的时间都很长，这样是无法做到极短时间内预测较长视频序列。所以我们提出了一个多帧预测的流程，可以同时提取不同时间点视频帧的特征，并且输出时也是同时输出预测T帧，通过这种方式就可以极大缩短预测的时间。

\vskip 0.1in \noindent\textbf{Transformer} 
%\label{subsec:transformer}
has been utilized extensively in NLP due to its benefits over RNN in feature extraction and long-range feature capture. It monitors global attention to prevent the loss of prior knowledge which often occurs with RNN. Its parallel processing capacity can significantly accelerate the process.
Recently, the field of computer vision has begun to explore its potential and produced positive results~\cite{dosovitskiy2020image,liu2021swin,liu2021video,arnab2021vivit,liang2021swinir}. Most related works input segmented patches of images to the transformer to calculate inter-patch attention and obtain the features.
There are also a number of vision transformer (VIT) approaches applied to video analysis. For example, VIVIT~\cite{arnab2021vivit} proposes four different video transformer structures to solve video classification problems using the spatio-temporal attention mechanism. \cite{liu2022video} applies the swin transformer structure to video and uses an inductive bias of locality. In this paper we select CNN as the feature extractor instead of the VIT structure because of the huge computational cost of VIT compared to CNN. We select the transformer structure as the predictor because it outperformed RNN, mix-mlp, and other structures, in terms of predicting spatio-temporal features in our empirical experiments.
\vskip 0.1in Most of the aforementioned video prediction methods extract from each frame complex features, typically of tens of thousands of bytes~\cite{shi2015convolutional,wang2018eidetic,guen2020disentangling,akan2021slamp}, resulting in excessive numbers of floating point operations in both the feature extraction module and the prediction module. Moreover, they employ sequential (frame-by-frame) prediction process. Hence, both training and testing consume a great deal of time and memory. In the meanwhile, many videos, particularly human activity records, have a significant amount of background redundancy~\cite{schuldt2004recognizing,h36m_pami} that can be removed by extracting information only from the key motions. 
Therefore, in this work, we try to couple the unique advantages of the transformer and the keypoint-based prediction methods to maximize their benefits.
%
%Note that the commonly adopted feature extraction based methods normally involve complex data distribution, for which transformer requires massive training data and well-trained or pre-trained model. However, keypoint prediction is a much easier and faster task for transformer thanks to its simple and sparse data distribution.
\begin{figure}[t]  
\centering
\includegraphics[width=10cm]{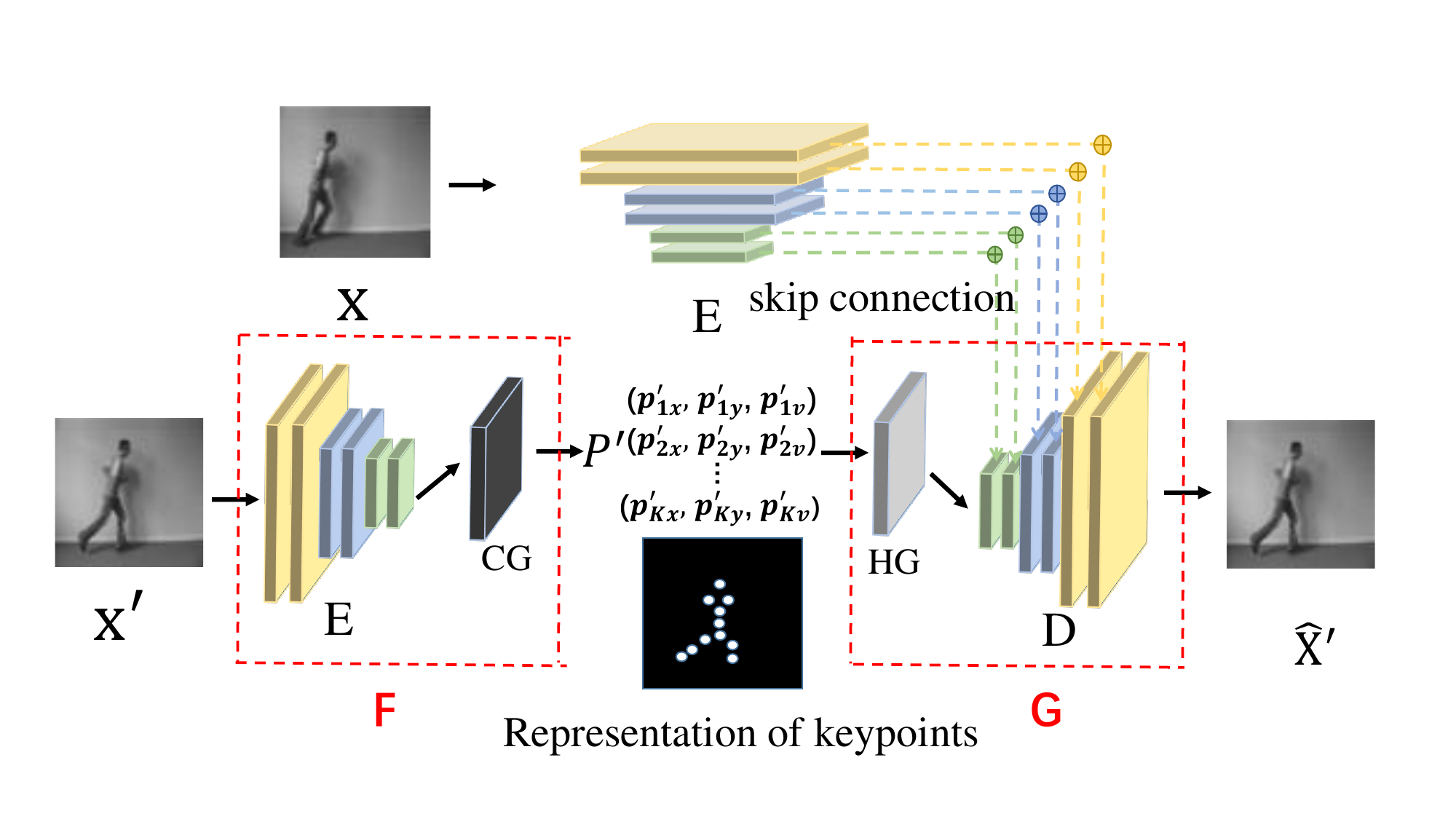} 
%\vspace{-0.3in}
\caption{Detailed structure of Keypoint Detector}       %对图进行说明 
\label{structure-detector}  
%\vspace{-0.1in}
\end{figure}

\begin{figure} [t]
\centering
\subfigure[]{
\begin{minipage}[t]{0.3\linewidth}
\centering
\includegraphics[width=0.64\textwidth]{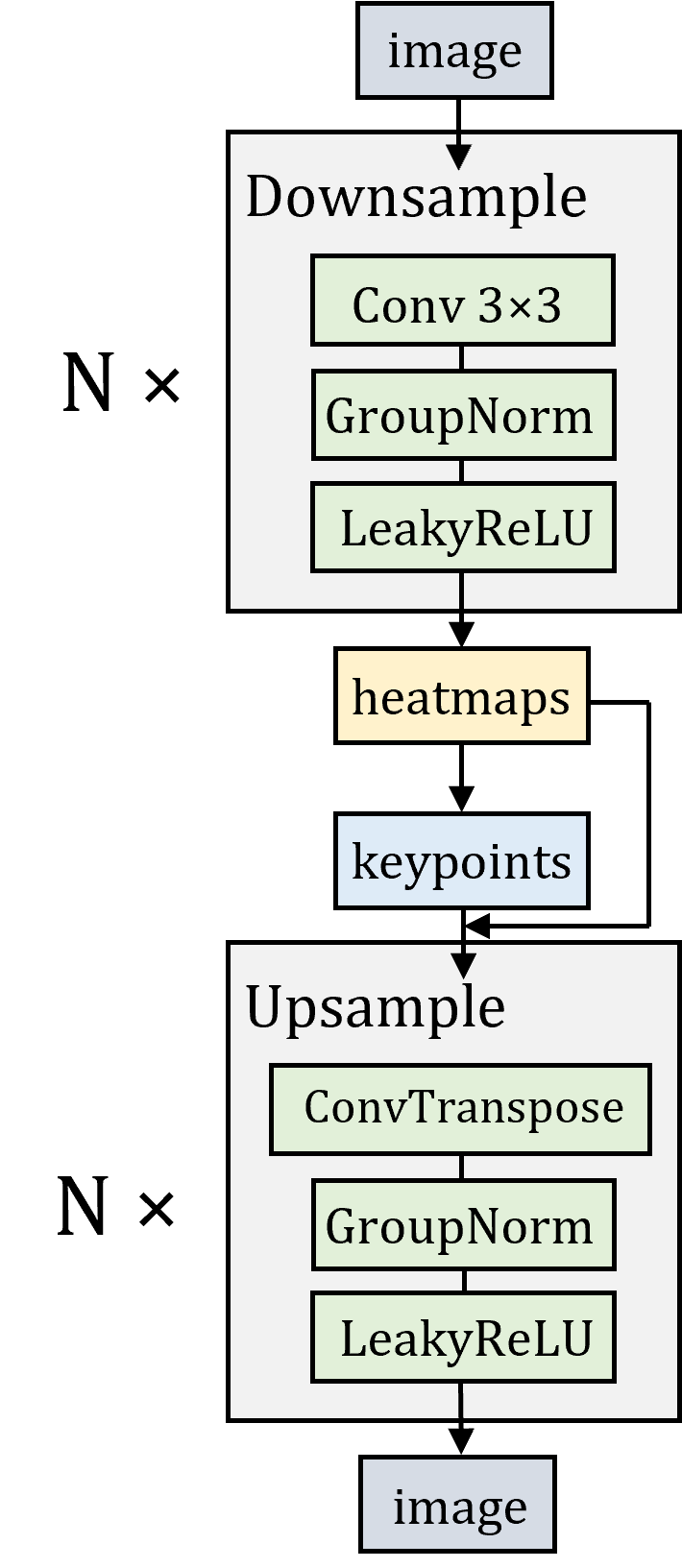}
%\caption{fig1}
\label{fig:encoder1}
\end{minipage}%
}
\subfigure[]{
\begin{minipage}[t]{0.3\linewidth}
\centering
\includegraphics[width=1.1\textwidth]{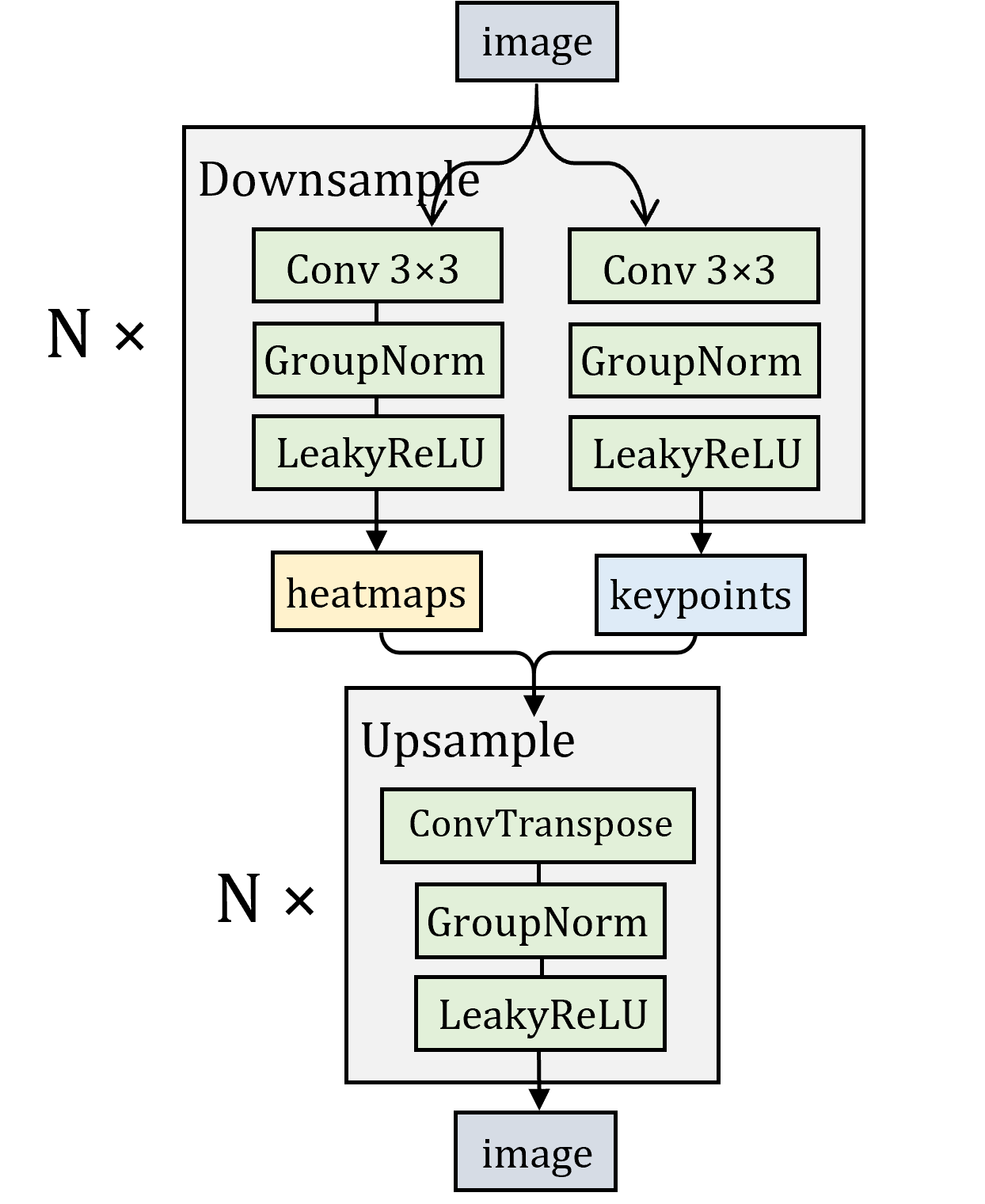}
%\caption{fig1}
\label{fig:encoder2}
\end{minipage}%
}
\subfigure[]{
\begin{minipage}[t]{0.3\linewidth}
\centering
\includegraphics[width=0.8\textwidth]{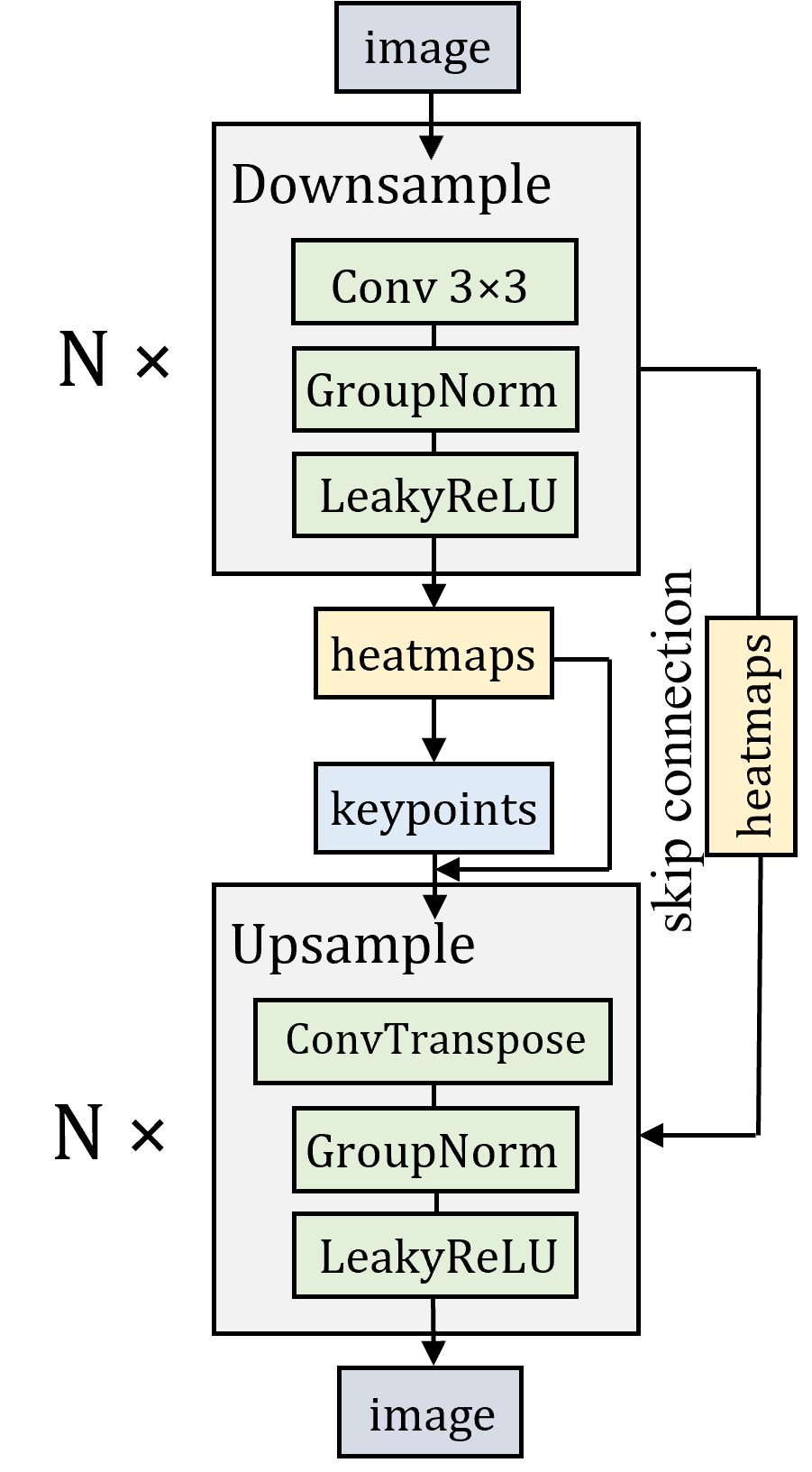}
%\caption{fig1}
\label{fig:encoder3}
\end{minipage}%
}

\caption{ Comparison of three different encoder and decoder structures. (a) The structure proposed by \cite{minderer2019unsupervised}  requires more network layers while performing poorly at disentangling keypoints and background information. (b) A structure that can well disentangle keypoints and background information at the cost of complex network architecture and high computation cost. (c) We adopt the well-known skip connection to achieve good performance on information disentangling with simple structure.}
	\label{fig:encoder_structure} 
\end{figure}

\begin{figure*}[t]  
\centering
\includegraphics[width=15cm]{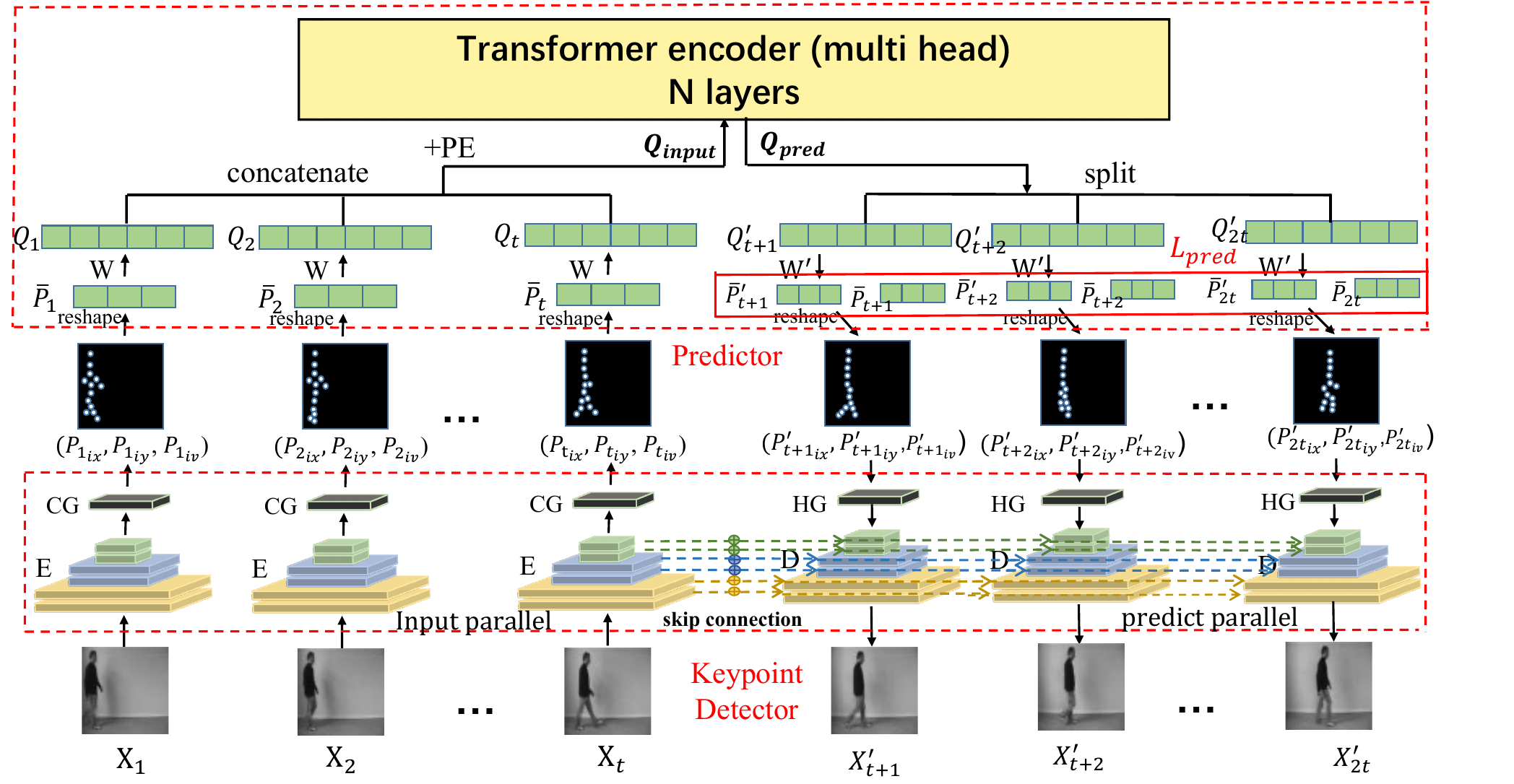} 
%\vspace{-0.15in}
\caption{Detailed structure of TKN. Two main modules are the Keypoint Detector and the Predictor marked with the red dashed lines. The predicted frame uses the background information extracted from the last frame of the input. Both the inputting stage and prediction stage allow batch processing (e.g., input multiple frames simultaneously) and thus enable temporal parallelism. Note that the ground truth keypoints information, $P_{real}=(\bar{P}_{t+1},...,\bar{P}_{2t})$, is output by $X_{t+1},...,X_{2t}$ using keypoint detector (excluded from the figure for simplicity).}       %对图进行说明 
\label{structure-1}  
%\vspace{-0.1in}
\end{figure*}

\section{Model}\label{sec:model}

%We first formally define the problem of video prediction as follows. Given a video stream consisting of continuous video frames, we denote by $\boldsymbol{X}=(X_{t-n+1},...,X_{t-1},X_t)$ the existing $\emph{n}$ video frames, where $ X_i \in \mathbb{R} ^{H \times W\times C} $ is the $\emph{i}$-th frame, $\emph{H}$, $\emph{W}$ and $\emph{C}$ are the height, width and number of channels, respectively. 
We start by formally defining the video prediction problem as follows. Given a stream of $\emph{n}$ continuous video frames, $\boldsymbol{X}=(X_{t-n+1},...,X_{t-1},X_t)$, $ X_i \in \mathbb{R} ^{H \times W\times C} $ denotes the $\emph{i}$-th frame for which $\emph{H}$, $\emph{W}$, and $\emph{C}$ denote the height, width, and number of channels, respectively. 
The objective is to predict the next $m$ video frames
$\boldsymbol{Y}=(Y_{t+1},Y_{t+2},...,Y_{t+m})$ using the input
$\boldsymbol{X}$.
\begin{equation}
 (X_{t-n+1},...,X_{t-1},X_t)\stackrel{predict}{\longrightarrow}
    (Y_{t+1},Y_{t+2},...,Y_{t+m}).
\end{equation}
Next, we present the dedicated design of \sysname\ for this
task. As depicted in Figure \ref{structure-1}, \sysname\ consists of two main modules, namely, the keypoint detector and the predictor module. 

\subsection{Keypoint Detector}
\label{subsec:detector}
\sysname\ employs a keypoint detector to detect the keypoints that are most likely moving. As illustrated in Figure \ref{structure-detector}, the detector extracts the keypoints as coordinate points. %\textcolor{blue}{Due to space limitations, we only introduce here the abstract representation, the training process and the structure of the encoder and decoder, other details we put in the Supplementary Material.}
Here we describe the abstract representation, training procedure, and the structures of the encoder and the decoder. Please refer to the supplemental material for additional information. 

\vskip 0.1in \noindent\textbf{Abstract representation.}Let $ X,X^{'} \in \boldsymbol{X} $ denote any two frames in $\boldsymbol{X}$, and $X$ referred as the source frame and $X^{'}$
the target frame. The keypoints in a video frame can be represented by ${P}=(p_{1},p_{2},...,p_{K}) \in {\Omega}^K$, where $K$ represents the number of keypoints, $ {\Omega}$ the coordinates. Assume that function $\mathbb{F}$ can extract the keypoints and $\mathbb{G}$ can reconstruct the target frame $X^{'}$ by using $K$ keypoints of $X^{'}$ and the features of the source frame $X$ :
\begin{subequations}
\begin{numcases}{}
\mathbb{F}(X^{'}) = P^{'}   \label{keydete}\\
\mathbb{G}(X;P^{'}) = \hat{X^{'}} \label{reconstrct},
\end{numcases}
\end{subequations}
where $\hat{X^{'}}$ denotes the reconstructed frame. By minimizing the difference between $\hat{X^{'}}$ and $X^{'}$ , the $P^{'}$ obtained by $\mathbb{F}$ represents the different parts between $X$ and $X^{'}$, which become what we call \emph{keypoints}. We use the pixel-wise $L_2$ frame loss to measure the difference between ${X^{'}}$ and $\hat{X^{'}}$ as follows:
\begin{equation}
 L_{rec}=  \Vert X^{'}-\hat{X^{'}} \Vert_2 \label{recloss}.
\end{equation}
%
%\pz{the superscript "2" should be removed? what's "rec"?}  
%It can be found that we do not label $X$ and $X^{'}$, and can learn both fuction $\mathbb{F}$  and $\mathbb{G}$  just using $ L_{rec}$ in \eqref{recloss} so  it is  an end-to-end unsupervised learning process.
$\mathbb{F}$ and $\mathbb{G}$ can be learned using $ L_{rec}$ in an end-to-end unsupervised learning process without labeling $X$ or $X^{'}$.

As shown in Figure \ref{structure-detector}, $\mathbb{F}$ consists of a n-layer CNN encoder $E$, and a coordinate generator $CG$ which converts each heatmap output of $E$ to $p_i^{'}=(p_{ix}^{'},p_{iy}^{'},p_{iv}^{'})$, where $p_i^{'}$ denotes the $i$-th keypoint of $X^{'}$, $(p_{ix}^{'},p_{iy}^{'})$ represents the coordinates of $p_i^{'}$ and $p_{iv}^{'}$ denotes the intensity. $\mathbb{G}$ consists of a heatmap generator $HG$ which converts the $K$ keypoints to a heatmap and a n-layer CNN decoder $D$ which has a symmetrical structure with $E$. %Please refer to the supplementary material for the detailed structures of $CG$ and $HG$.
\vskip 0.1in \noindent\textbf{Encoder and Decoder.} %\textcolor{red}{the paragraph title is encoder but the paragraph also introduces the decoder, some content is overlapped with "Decoder" paragraph and should be removed}
%\textcolor{blue}{ It is clearer to present the encoder and decoder together when introducing the structure of Figure 3}
We compared three structures of the encoder and the decoder. The first structure, as shown in Fig.~\ref{fig:encoder1}, is proposed by Minderer~\etal~\cite{minderer2019unsupervised}, in which the output heatmaps serves both as the generation feature of keypoints and the input feature for the background of the decoder (the heatmaps here corresponds to X in Eq.~\eqref{reconstrct}). Although this structure is simple, it needs a lot of encoder layers and a high feature dimension to extract both key points and background information. Besides, our experimental results show that this structure does not perform well in reconstructing the target frame. The structure in Fig.~\ref{fig:encoder2} uses two networks to extract the keypoints information and the background information separately. While it performs well at information disentangling in our experiments, its structure is too complex and thus requires high computation cost. Therefore, in \sysname\, we design a structure shown in Fig.~\ref{fig:encoder3} adopting the ``skip connection'' proposed by~\cite{ronneberger2015u}, to allow the encoder to disentangle the background information layer by layer and only focus on outputting the keypoints, and synthesize the disentangled background information into the decoder via skip connection. Experimental results show that the structure in Fig.~\ref{fig:encoder3} can reconstruct frames better with a lower computation cost.%, so we choose the structure in Fig.~\ref{fig:encoder3} for the encoder and decoder. \textcolor{red}{Fig.~\ref{fig:encoder2} is not used?  Why keep it? Maybe better remove it.}\textcolor{blue}{The structure in Figure 3(b) is the simplest way to separate keypoints and  background information, and serves as a bridge between the structure Figure 3(a) and Figure 3(c) , as well as providing a comparative experiment for c, showing that the  structure of Figure 3(c) is  simpler and can separate the information at the same time}
\par Let $E_i$ denote the $i$-th layer of $E$ and $h_i \in \mathbb{R} ^{H_i \times W_i\times C_i} $ denote its output heatmap, where $h_1^{'}=E_1(X^{'})$ and each subsequent layer can be expressed as follows:
\begin{equation}
E_{i}(h_{i-1}^{'}) = h_{i}^{'}, i\in\{2,n\}.
\end{equation}  

\par
The outout of the last layer $h_n^{'}$ is fed into coordinate generator $CG$ to extract the keypoints :
\begin{equation}
CG(h_n^{'}) =(p_1^{'},...,p_i^{'},...,p_K^{'}).
\end{equation} 
%where $p_i^{'}=(p_{ix}^{'},p_{iy}^{'},p_{iv}^{'})$ and these $K$ keypoints  correspond to $P^{'}$ in the previous section.
\vskip 0.1in \noindent\textbf{Coordinate Generation (CG)}
module converts the heatmap generated by the encoder's last layer to the keypoints. % in the keypoint detector .
%The role of CG is to convert the heatmap into coordinate form.
We use a similar CG structure as in \cite{jakab2018conditional} which first uses a fully connected layer to convert the encoder heatmap $h_n$ from $ \mathbb{R} ^{H_n \times W_n\times C_n}$ into $ \mathbb{R} ^{H_n \times W_n\times K}$, where $K$ refers to the number of keypoints. We do this in the hope of compressing $H_n \times W_n$ into the form of point coordinates in the dimension of $K$. The converted heatmap $h^{'}_n$ can be rewritten as $h^{'}(x;y;i)$, where $x=1,2,...,W_n,y=1,2...,H_n,i=1,2,...,K$, represent the three dimensions of $h^{'}_n $, respectively. Then we can calculate the coordinates of the $k$-th keypiont in width $ p_{ix}$ as follows:
\begin{equation}
h_n^{'}(x;i)= \frac{\sum_{y} h_n^{'}(x;y;i)}{\sum_{x,y} h_n^{'}(x;y;i)}  , \label{4}
\end{equation}
\begin{equation}
p_{ix}= \sum_{x} h_{x}h_n^{'}(x;i)   ,  \label{5}     
\end{equation}
where $h_x$ is a vector of length $W_n$ consisting of values uniformly sampled from -1 to 1 (for example , if $W_n=4$ then $h_x=[-1,-0.333,0.333,1]$). By doing so, we add an axis to the heatmaps at the dimension $W_n$ where $p_{ix}$ is the position of the $i$-th keypoints on the width. Similarly, we can calculate the coordinate in height $ p_{iy} $ by exchanging the position of $x$ and $y$ using Equation~\eqref{4} and Equation~\eqref{5}. We also need the feature values at these coordinates to reconstruct the following frames with keypoints. We express such values with the averages on both the $H_n$ and $W_n$ dimension. We use $p_{iv}$ to represent the value of the $k$-th keypoint:
\begin{equation}
   p_{iv}=\dfrac{1}{H_n \times W_n} \sum_{x,y}h_n(x;y;i).
\end{equation}
As such , we extract the keypoints as $p_i=(p_{ix},p_{iy},p_{iv})$ ,$i=1,2,...,K$.

\vskip 0.1in \noindent
\par Next, the features of $X$ and $P^{'}$ are input to $\mathbb{G}$ as shown in  Eq. \eqref{reconstrct}. The features of $X$, $h_n \in  \mathbb{R} ^{H_n \times W_n\times C_n}$, are obtained via $E$. $P^{'}$ is converted to a heatmap $h_p^{'}$ via $HG$ :
\begin{equation}
HG(p_1^{'},...,p_i^{'},...,p_K^{'})=h_p^{'}.
\end{equation}

\vskip 0.1in \noindent\textbf{Heatmap Generation (HG)}
module is a reversed process of CG that converts coordinates to the heatmap. We use a 2-D Gaussian distribution to reconstruct the heatmaps. We first convert the coordinates $p_x,p_y$ into 1-D Gaussian vectors, $x_{vec}$ and $y_{vec}$, where $p_x=(p_{1x},p_{2x},...,p_{Kx}),p_y=(p_{1y},p_{2y},...,p_{Ky})$, as follows:
\begin{equation}
x_{vec}=exp(-\frac{1}{2\sigma_{2}}\Vert p_x - \bar{p}_x \Vert^{2}),
\end{equation}
\begin{equation}
y_{vec}=exp(-\frac{1}{2\sigma_{2}}\Vert p_y - \bar{p}_y \Vert^{2}),
\end{equation}
where $\bar{p}_x$ and $\bar{p}_y$ are the expectations of $p_x$ and $p_y$, respectively. By multiplying $x_{vec}$ and $y_{vec}$ we can get the 2-D Gaussian maps $G\_maps$ as follows:
 \begin{equation}
    G\_maps=x_{vec} \times y_{vec}.
 \end{equation}
Finally we calculate the Hadamard product of $G\_maps$ and $p_v$ to get the $h^{'}_p$:
  \begin{equation}
    h^{'}_p=G\_maps \circ p_v .
 \end{equation}
We align the dimension of $h_p^{'}$ with $h_n$ to allow their direct concatenation sent to the decoder for reconstruction.
As mentioned, inspired by the ``skip connetion'' in UNet~\cite{ronneberger2015u} and Ladder Net~\cite{rasmus2015semi}, which can reconstruct images better with fewer encoder and decoder layers, we input the heatmaps $h_1,h_2,..h_n$ obtained by each encoder layer to the decoder through ``skip connection''. Let $D_i$ denote the $i$-th decoder layer and $d_i$ denote its output heatmap, where  $d_1=D_1(concat(h_p^{'},h_n))$ and each subsequent layer can be expressed as follows:
\begin{equation}
d_{i}=D_{i}(concat(d_{i-1},h_{n-i+1})), i \in \{2,n\} ,
\end{equation}
where $d_n$ is $\hat{X^{'}}$ . In this manner, the decoder learns the ``background'' (i.e., the static information) features eliminated by the encoder, thus improving the higher level representation details of the model. The additional ``background'' information also allows the encoder to focus more on the keypoints. %\textcolor{blue}{Compared with~\cite{minderer2019unsupervised}, \sysname\ has fewer encoder and decoder layers and requires fewer keypoints for frame reconstruction.}

% \begin{figure}[t!]
% \centering
% \includegraphics[width=4cm]{figures/transformer.png}
% \caption{The structure of transformer encoder. Embedded keypoints are the explicit keypoints coordinates mapping to the high-dimensional implicit representations combined with temporal position encoding.}
% \label{fig:transformer}
% \end{figure}
\subsection{Predictor}

The original prediction task is transformed, via the keypoint detector's encoding, into predicting the subsequent $m$  groups of keypoints $P_{t+1},...,P_{t+m}$, based on the prior $n$ groups of keypoints $P_{t-n+1},...,P_{t}$. We select the transformer\cite{vaswani2017attention}as the predictor because it can associate keypoint information at each moment through attention, making it less prone to forgetting compared to sequential networks like RNNs. We found that using only the transformer's encoder for encoding temporal relationships between keypoints leads to better and faster predictions than using the entire structure.

\begin{figure} [t]
\centering
\subfigure[]{
\begin{minipage}[t]{0.3\linewidth}
\centering
\includegraphics[width=1\textwidth]{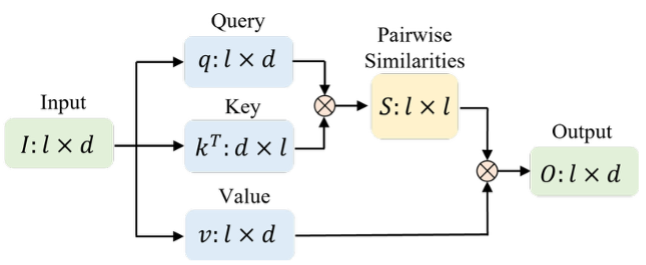}
%\caption{fig1}
\end{minipage}%
}
\subfigure[]{
\begin{minipage}[t]{0.3\linewidth}
\centering
\includegraphics[width=1\textwidth]{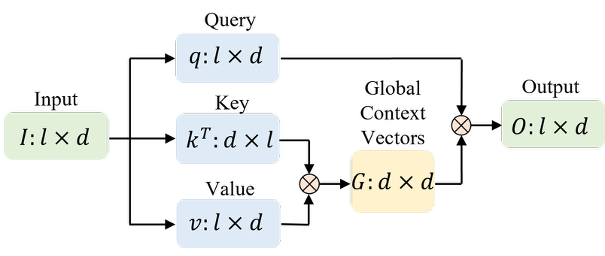}
%\caption{fig1}
\end{minipage}%
}
\subfigure[]{
\begin{minipage}[t]{0.3\linewidth}
\centering
\includegraphics[width=1\textwidth]{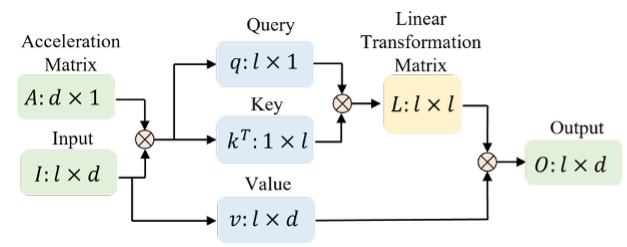}
%\caption{fig1}
\end{minipage}%
}

\caption{ Three different attention mechanism structures: (a) is the original dot-product structure, with a  computational complexity of $O(l^2 d)$; (b) is the linear attention structure, with a computational complexity of $O(ld^2)$; (c) is the structure we propose, which uses an acceleration matrix A to reduce the computational complexity of the linear transformation matrix L to $O(l(d+l)^2)$.}
	\label{fig:attention_structure} 
\end{figure}

Transformer uses the attention mechanism, utilizing query ($q$), key ($k$), and value ($v$) to compute the correlation between sequence nodes. The computational complexity of this attention is $O(l^2 d)$ ($l$ represents the sequence length, and d is the dimension of the sequence) as shown in Fig.\ref{fig:attention_structure}(a). Many methods aim to reduce attention's computational complexity, such as linear attention, which assumes $l>>d$ in natural language processing and results in a complexity of $O(ld^2)$ as shown in Fig.\ref{fig:attention_structure} (b). However, in video prediction, l<d, so such improvements would increase computational complexity. To address this, we introduce an acceleration matrix $A\in R^{d_k×1}$  , which, when multiplied with the input I, yields $q_A$ and $k_A$ matrices with reduced dimensions. We then compute the linear transformation matrix $L=q_A k_A^T$ to reduce the computational complexity to $O(ld+l^2)$ as shown in Fig.\ref{fig:attention_structure} (c). This reduces the overall complexity by half. We only apply this operation to the $qk^T$ matrix to maintain prediction accuracy.

The transformer encoder requires the input to be a one-dimensional vector with a length of $d_model$, e.g., (512,768,1024). Hence we first convert the K keypoint-triples $\{(p_ix,p_iy,p_iv)|i=1,..,K\}$ to a one-dimensional vector $\bar{P}$ :
\begin{equation}
    \bar{P} = (p_{ix},p_{iy}, p_{iz},...,p_{Kx},p_{Ky},p_{Kz})
\end{equation}

\begin{figure*}[t!]
\centering
\includegraphics[width=\textwidth]{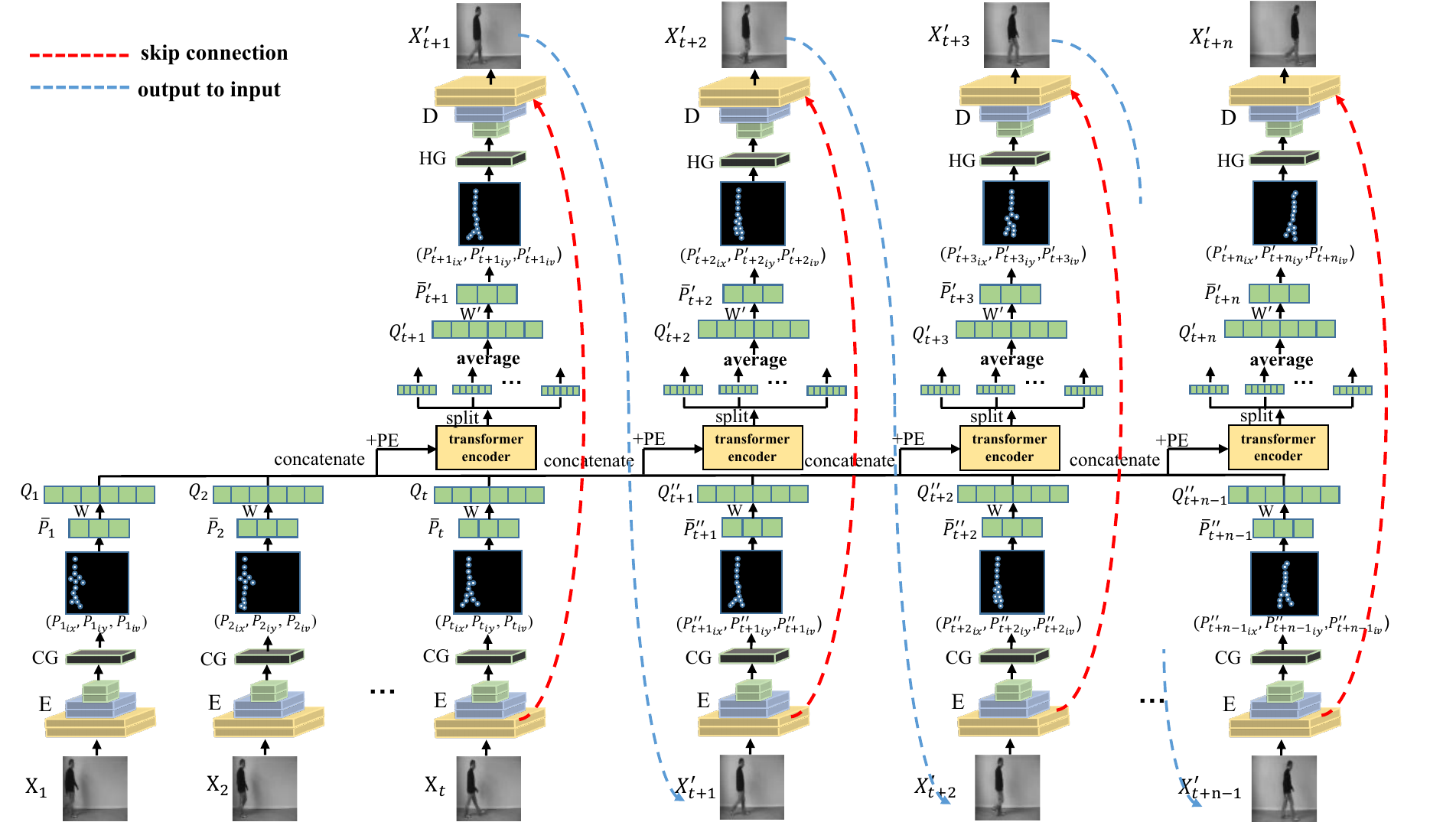}
\caption{Detailed structure of TKN-Sequential. It uses the same keypoint detector and predictor structures with TKN but has a different prediction process. Particularly, it uses the previous predicted frame's background as the following one's to ensure background consistency.}
\label{fig:tkn-long2}
\end{figure*}

$\bar{P}$ represents the low-dimensional \textit{explicit} spatial coordinates. Inputting $\bar{P}$ directly into the transformer necessitates adjusting the dimension of the intermediate parameter values according to $K$ in each prediction instance. Consequently, training and testing would become difficult.
%而且P表示的是空间位置的低维显式坐标，目前已经有大量工作（比如nerf）表明了了高维隐示表达在处理复杂问题时的好处，所以我们基于这些问题，使用一层MLP将P转换位固定长度的高维隐示空间特征。

The predicted keypoints sequence is obtained through linear transformation of the input keypoints sequence. However, real-world object motion is often complex and better represented by differential equations. By transforming the explicit input into a higher-dimensional latent space, the keypoint sequence features can be predicted through linear transformation. Moreover, many works~\cite{meng2018zero,tao2020latent,zhou2021latent} have demonstrated the benefits of \textit{latent} representations. % facing complex problems\textcolor{blue}{
We also found latent representations well capture the regularity of keypoints over time in our experiments. %Unlike~\cite{mildenhall2021nerf,kasten2021layered,bar2022text2live} which use network parameters as implicit representations, 
Therefore, we use a matrix to map the explicit coordinates representation to a latent space to obtain a high-dimensional latent representation vector. Specifically, we convert the variable-length and low-dimensional $\bar{P}$ to a fixed-length and high-dimensional $Q$ by converting $\mathbb{R}^{3K}$ to $\mathbb{R}^{d_{model}}$ via a mapping matrix $W$: $Q = W \cdot \bar{P}$ . where $W \in \mathbb{R}^{d_{model} \times 3K }$. To compensate the lack of time-sensitive capability, we manually add location information position embedding (PE) to $Q$, 
\begin{equation}
   Q_{input} = concat(Q_{t-n+1},...,Q_{t}) + PE , \label{Qinput}
\end{equation} 
where PE is the trigonometric function as defined in \cite{vaswani2017attention}. The number of input sequences and output sequences are equal for the transformer encoder. Therefore, we can use transformer encoder to get the predictions $Q_{t+1}^{'},...,Q_{2t}^{'}$ using the input $Q_{input}$:
\begin{equation}
Q_{t+1}^{'},...,Q_{2t}^{'} = Trans\_encoder(Q_{input}).
\end{equation}

Finally, we use an invert mapping matrix $ W^{'} \in  \mathbb{R}^{3K \times d_{model}} $ to reconstruct the high-dimensional sequence $Q_{pred}=(Q_{t+1}^{'},...,Q_{2t}^{'})$ back to the low-dimensional keypoints spatio-temporal sequence $P_{pred}$, which is then input to the decoder to generate the predicted frames:
\begin{equation}
    P_{pred} = W^{'} \cdot Q_{pred}.
\end{equation}
We only need to calculate the loss of sequence $P_{real}=(\bar{P}_{t+1},...,\bar{P}_{2t})$ which is output by $X_{t+1},...,X_{2t}$ using keypoint detector and $P_{pred}$ to complete the training of the predictor using a well-trained keypoint detector, for which we also use the $L_2$ loss:
\begin{equation}
L_{pred}= \Vert P_{real} -P_{pred} \Vert_2  \label{predloss}.
\end{equation}

%We noticed via empirical evaluations that the performance of a single transformer encoder was limited regardless of the encoder depth and the number of heads. Therefore, we choose to use multiple encoders each of which predicts one output. This structure increases the prediction accuracy significantly with negligible reduction of the prediction speed. Let  $\boldsymbol{Q}=(Q_{t-n+1},...,Q_{t})$ denote the transformer input and $(Q_{t+1},Q_{t+2}..  Q_{t+m})$ the prediction target. Assuming we have successfully predicted $(Q_{t+1},...,\hat{Q}_{t+i}) $, the next $\hat{Q}_{t+i+1}$ can be denoted as:

%

\subsection{Prediction Processes}\label{subsec:prediction}
%我们之前讲了，我们把以前方法类似于循环网络结构的预测流程称为逐帧预测，我们的称之为多帧预测，我们并行提取特征是为了追求预测时的推理速度，追求速度方法往往会带来精确度方面的下降，所以我们将我们的主要结构不做变化，仅仅将多帧预测流程变成逐帧预测的流程，对比图如图所示。这两种流程最大的区别是，在多帧预测流程中，我们使用输入帧最后一帧的背景，作为其他预测帧的生成，而逐帧预测流程中，是用相邻帧的背景，作为下一帧的生成，这里需要注意即便是时间上非常近的两帧，提取到的背景信息也不是完全一样的，所以并不会出现每次用相同的背景信息作为预测帧的合成，否则就和多帧预测没区别了。
%\textcolor{red}{As we mentioned before, we call the prediction process of the previous method , which similar to the recurrent neural network structure frame-by-frame prediction, ours is called multi-frame prediction. We extract features in parallel in order to pursue the speed of inference when predicting, and usually the pursuit of speed methods often brings about a decrease in accuracy, so for comparison we leave our main structure unchanged and simply turn the multi-frame prediction process into a frame-by-frame prediction process, and the comparison is shown in \ref{fig:tkns}. It use the background information of the last frame of the input frames as the generation of the other prediction frames in the multi-frame prediction process . However in the frame-by-frame prediction process ,it uses the background information of a newly predicted frame as the following target's background.}
%\begin{comment}
    
Sequential prediction is time consuming. Since most subsequent frames in high frame-rate videos are fairly similar, we can use the background of the frame immediately just before the prediction target as the background of the prediction target frame. We can then combine the predicted keypoints with the background to generate the integrated predicted frames. As illustrated in Fig.~\ref{structure-1}, we integrate the $P_{pred}$ generated by Transformer encoder to the background of t-th frame and directly generate the subsequent t prediction frames $X_{t+1}^{'},X_{t+2}^{'},...,X_{2t}^{'}$. This parallel prediction mechanism, i.e., TKN, can input and predict multiple frames as batches to significantly accelerate the prediction process. %Note that the ``background'' in this context also refers to the parts that do not change between frames.

%Specifically, as Figure~\ref{fig:tkn} depicts, \emph{TKN} employs a multi-head encoder which converts prior $T$ keypoints $\{p_1,p_2,...,p_T\}$ to future $T$ keypoints $\{p_{T+1}, p_{T+2}, ..., p_{2T}\}$, and uses the background of frame $X_T$ as that of the frames $\{X_{T+1}, X_{T+2}, ..., X_{2T}\}$. 

We reason that frame-by-frame prediction structure has higher accuracy to predict frames with frequent changes (as proved by experiment results). Hence we also provide a sequential variation of TKN, TKN-Sequential, which uses the previous predicted frame's background as the following one's to ensure background consistency. Fig.~\ref{fig:tkn-long2} shows the detailed structure of \sysname-Sequential amd Fig.~\ref{fig:tkn-long2} depicts its comparison with TKN.

Since the output of the transformer encoder has the same length as the input sequence, we take the averaged predictor's output as the predicted frame. 
%\pz{the motivation is not clear, why output has same length with input justify using the avg?}. %\textcolor{red}{The output of each step of TKN Long is predicted one frame, but the transformer encoder is N input N output at a time, so you need to compress the N outputs into one output, and here the avg value is used to take one output} .
Suppose we have predicted $i$ frames, $Q_{t+1}^{'},...,Q_{t+i}^{'}$, then $ Q_{t+i+1}^{'}$ can be expressed as:
\begin{equation}
    Q_{t+i+1} ^{'} = \dfrac{1}{t+i}\sum_{j=1}^{t+i}Trans\_encoder(Q_{input};j) ,
\end{equation}
where $Q_{input}=concat(Q_{t-n+1},...,Q_{t},Q^{'}_{t+1},...,Q_{t+i}^{'}) + PE$ according to \eqref{Qinput}.

 $X^{'}_{t+1}$ is the combination of predicted $\bar{P}^{'}_{t+1}$ and the background information of $X_{t}$ extracted by the decoder. Note that  ``$\bar{P}^{''}_{t+1}$ and the background information of $X^{'}_{t+1}$'' are not equal to ``$\bar{P}^{'}_{t+1}$ and the background information of $X_{t}$'', albeit they are both extracted by the encoder from the $X^{'}_{t+1}$. It is because two consecutive frames are very similar but still have some minor differences in the keypoints and background information, otherwise it would be no different from multi-frame prediction process.

\section{Experimental Setup}

\label{sec:setup}
\vskip 0.1in \noindent\textbf{Dataset.}
\begin{table*}[t!]
\centering
 \resizebox{\textwidth}{40mm}{
\begin{tabular}{c|c|c|c|c|c|c|c|c}  \hline
         Dataset&Method& SSIM$\uparrow$  & PSNR $\uparrow$  & TIME (s)$\downarrow$& TIME (ms)$\downarrow$& FPS$\uparrow$ &Memory (MB)$\downarrow$ & Memory (MB)$\downarrow$ \\   
                 &    &      &      & (train)& (test)& (test)&(train) &(test)    \\         \cline{1-9}
     \multirow{9}{*}{KTH}& ConvLSTM &0.712*  & 23.58* &   61 & 72&278&    8,055 & 1,779             \\ 
      &  PredRNN & 0.839*  & 27.55* &   204 & 184  &109&    6,477& 1,721              \\  
     &  PredRNNv2&0.838*  &  28.37* &   246 &  222 &    90&8,307  & 1,779              \\
      & PhyDNet& 0.854  & 26.9  &   108  &  240 &83&    8,491 &  2,704             \\
      &  SLAMP&     0.864*(30)  &   28.72(30)  &465 & 388&52&   21,103(16)  &  2,295   \\  
    &    E3D-LSTM&  \textbf{0.879}* & \textbf{29.31}* &   879 &   338 &59&   21,723(16)  & 2,687              \\  \cline{2-9}
    & Grid-Keypoint&  0.837* & 27.11*  & 145 &  252 & 79&  12,661 & 2,259              \\ 
     & Struct-VRNN&  0.766* & 24.29* & 111 & 151  &  132& 5,661  &   1,817            \\  \cline{2-9}
   &    \textbf{TKN (w/o tp)}  &0.871 & 27.71  & \textbf{35}  &  86  &  233 & \textbf{ 3,777}  & \textbf{1,447}              \\   
   &    \textbf{TKN-Sequential} &  0.862 & 27.73 & 44  &   154  &130  &  6,309  &  1,785         \\ 
   &    \textbf{TKN}  &0.871 & 27.71 & \textbf{35}  &  \textbf{17}  &  \textbf{1,176} & 4,945  & 1,705              \\\hline
   \multirow{7}{*}{Human3.6}& ConvLSTM &   0.776*& - &  63      &  32 & 125 &  6,561&   1,857        \\  
   &    PredRNN &    0.781* & - & 462     &   47&  85 &   5,829   &  1,743             \\   
    &   E3D-LSTM &    0.869*  &  - &  3154 &    167&  24&    18,819(8)    &  5,767             \\   
    &   PhyDNet &    0.901* &  -& 207  &  88   & 45 &      12,213&  2,353             \\     \cline{2-9}
    &  Grid-Keypoint &   0.928   & 28.76&  114  &  106& 38 &      9,891     &2,003      \\
& Struct-VRNN&  0.916  &  26.97 &   67  &  41   &   98    &   5,015     &    1,962      \\   \cline{2-9}
  &     \textbf{TKN (w/o tp)} &  \textbf{0.958}  &  \textbf{30.89}   &    \textbf{63}&   30  & 133&  \textbf{2,179}    & \textbf{1,521}              \\   
  &   \textbf{TKN-Sequential} &   0.946 & 29.56  &    75 & 35 &114  &     2,653          &   1,763            \\  
   &\textbf{TKN} &  \textbf{0.958}  &  \textbf{30.89}   &    64&   \textbf{11}  & \textbf{364} &  2,561    & 1,587              \\   \hline
\end{tabular}
}
\caption{The results on KTH and Huamn3.6. $\uparrow$ means the higher the better and $\downarrow$ means the less the better. We skipped some tests due to the lack of original code. Instead, we used the results provided by the original papers (indicated by ``*''), or skipped if the papers didn't provide results (indicated by ``-''). ``w/o tp'' means without temporal parallel and ``seq'' means sequential. We used 32 and 16 as the default batch sizes for KTH and Human3.6, but 16 and 8 for a few exceptions which otherwise exceeded the GPU capacity due to too many intermediate results generated by the algorithms. (30) indicates using 10 input frames to predict 30 frames with SLAMP. %\pz{why (30)}.  \textcolor{red}{ Because the results of their code run on git are very poor, we quote the results from their paper,Only 10 predictions of 30 in his paper}. 
\textbf{Struct-VRNN} and \textbf{Grid-Keypoint} are Keypoint-based baselines.}
\label{tab:1}
\end{table*}
We used two real action datasets, KTH~\cite{schuldt2004recognizing} and Human3.6~\cite{h36m_pami}, to verify the real-time and efficient performance of the proposal under different patterns.
\begin{comment}

\par KTH dataset includes 6 types of movements (walking, jogging, running, boxing, hand waving, and hand clapping) performed by 25 people in 4 different scenarios, for a total of 2391 video samples. The database contains scale variations, clothing variations, and lighting variations. We use people 1–16 for training and 17-25 for testing. Each image is converted to the shape of $(64, 64, 3)$.
\par Human3.6 dataset contains 3.6 million 3D human poses performed by 11 professional actors in 17 scenarios (discussion, smoking, taking photos and so on). We use scenario 1, 5, 6, 7, and 8 for training and 9 and 11 for testing. Each image is converted to the shape of $(128, 128, 3)$.
\end{comment}

\begin{itemize}
    \item KTH dataset includes 6 types of movements (walking, jogging, running, boxing, hand waving, and hand clapping) performed by 25 people in 4 different scenarios, for a total of 2391 video samples. The database contains scale variations, clothing variations, and lighting variations. We use people 1–16 for training and 17-25 for testing. Each image is converted to the shape of $(64, 64, 3)$.
    \item Human3.6 dataset contains 3.6 million 3D human poses performed by 11 professional actors in 17 scenarios (discussion, smoking, taking photos and so on). We use scenario 1, 5, 6, 7, and 8 for training and 9 and 11 for testing. Each image is converted to the shape of $(128, 128, 3)$.
\end{itemize} 
%Please refer to the supplementary material for the details of dataset processing.
%\par We put the processing of the dataset in the Supplementary Material.
\vskip 0.1in \noindent\textbf{Implementation.}
The experiments were run on a server equipped with an Nvidia GeForce RTX 3090 GPU. 
We conducted a two-step training: first we trained the keypoint detector using $L_{rec}$ in Eq. \eqref{recloss} and then froze its parameters, then we trained the predictor using $L_{pred}$ in \eqref{predloss}. We found that this method trained faster than the end-to-end training which trained the keypoint detector and predictor together using both $L_{rec}$ and $L_{pred}$. %Please refer to the supplementary material for the detailed experimental structure.
%We used the loss image in keypoint detector training, and the loss image plus the loss keypoints in end-to-end training, with the $\lambda$ taken as 0.1. We put the detailed experimental structure in the supplementary files.
\vskip 0.1in \noindent\textbf{Model structures.}
TKN and TKN-Sequential have the same keypoint detector structure, which has a 6-layer encoder and a 6-layer decoder. Each encoder layer includes Conv2D, GroupNorm, and LeakyRelu. Each decoder layer includes TransposedConv2D, GroupNorm and LeakyRelu. Since the skip connection is used between encoder and decoder, the input dimension of each decoder layer is twice the output dimension of the corresponding encoder layer.
\par 
For the predictor, TKN uses a 6-layer transformer encoder with the input sequence length of 10, and TKN-Sequential uses 10 single-layer transformer encoders, each with an input sequence length of 10, 11, ..., 19. As mentioned in Section Prediction Processes, each transformer encoder's output of TKN-Sequential is averaged according to the input length, hence the output of both TKN and TKN-Sequential has a length of 10. All transformer encoders employed by the baselines share the same parameters: $d_k=d_v=64,d_{model}=512,d_{inner}=2048,n_{head}=8,dropout=0$
\vskip 0.1in \noindent\textbf{Evaluation metrics.}
Traditional evaluation metrics include structural similarity~(SSIM) and peak signal to noise ratio~(PSNR).
Higher SSIM indicates a higher similarity between the predicted image and the real image. Higher PSNR indicates better quality of the reconstructed image. 
We also quantify the resource (time and memory) consumption, for which a uniform batch size of 32 and 1 are used for KTH dataset during training and testing, and 16 and 1 are used for Human3.6 dataset during training and testing.

In addition, we measure the FLOPs (floating-point operations per second) to assess the computational cost and the number of parameters of the model with the thop\footnote{\url{https://pypi.org/project/thop/}} package. %\textcolor{red}{R1C3}

\vskip 0.1in \noindent\textbf{Baselines.}
To validate the performance of TKN, we select 8 most classical and effective SOTA methods as the baselines, all of which are implemented with Pytorch for fair comparisons.
The 8 baselines include:  ConvLSTM\cite{shi2015convolutional},  Struct-VRNN\cite{minderer2019unsupervised}, Grid-Keypoint \cite{gao2021accurate}, Predrnn\cite{wang2017predrnn}, Predennv2\cite{wang2022predrnn}, PhyDNet\cite{guen2020disentangling}, E3D-LSTM\cite{wang2018eidetic}, SLAMP\cite{akan2021slamp}.
\begin{enumerate}
\item  ConvLSTM\cite{shi2015convolutional} is one of the oldest and most classic video prediction method based on LSTM. 
\item Struct-VRNN\cite{minderer2019unsupervised} is the first one to use keypoints to make prediction. 
\item Grid-Keypoint\cite{gao2021accurate} is a grid-based keypoint video prediction method.
\item Predrnn\cite{wang2017predrnn} is a classic prediction method adapted from LSTM.
\item Predennv2\cite{wang2022predrnn} can be generalized to most predictive learning scenarios by improving PredRNN with a new curriculum learning strategy. %We can not train it on Human3.6, so we  only use it on the KTH dataset.
\item PhyDNet\cite{guen2020disentangling} disentangles the dynamic objects and the static background in the video frames. % that achieves the best performance on the Human3.6.
\item E3D-LSTM\cite{wang2018eidetic} combines 3DCNN and LSTM to improve prediction performance. % a prediction method combining 3DCNN and LSTM and is one of the most effective methods on KTH dataset. % one of the most effective methods on KTH dataset.
\item SLAMP\cite{akan2021slamp} is an advanced stochastic video prediction method. 
\item To highlight the importance of parallel prediction in terms of the fast prediction of TKN, particularly when compared with sequential keypoints method Struct-VRNN and Grid-Keypoint, we tested \emph{TKN(w/o tp)} which has the same structure as TKN but lacks the parallel scheme of the keypoint detector. 
\end{enumerate}
Due to the lack or incompleteness of open-sourced code, we tested Predennv2 and SLAMP only on KTH dataset while the others on both datasets.

\section{Results and Analysis}
\begin{figure*}[t!]
\centering
\includegraphics[width=\textwidth]{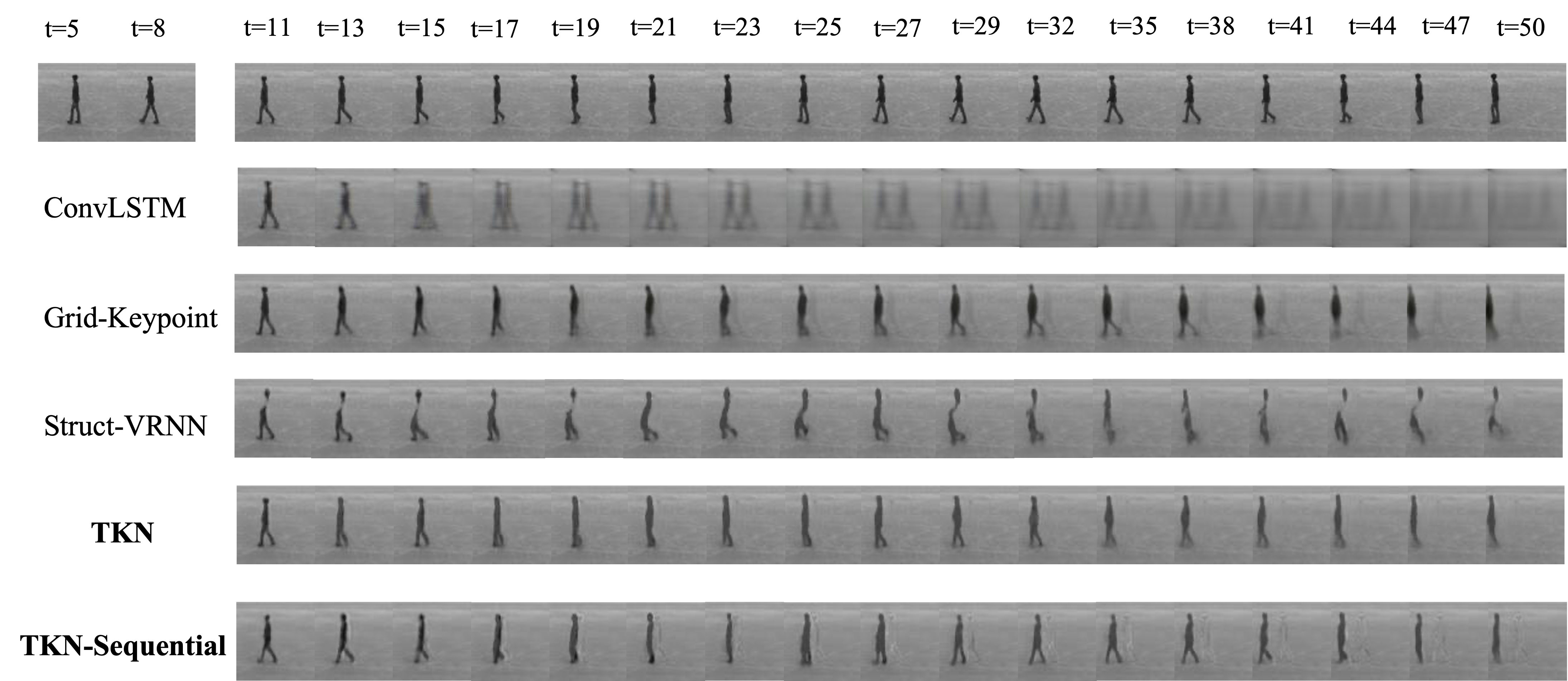}
\caption{Results of long-range predictions on KTH. TKN and TKN-Sequential perform better than the baselines. TKN-Sequential provides more precise details.}
\label{fig:kthlong}
\end{figure*}

\begin{table}[t]
 %   \vspace{-0.15in}
\centering
\setlength{\tabcolsep}{7mm}{
    \begin{tabular}{c|c|c|c}  \hline
            Method  &boxing&handclapping&handwaving \\ \hline
         TKN&  0.897   &0.908& 0.898      \\
         TKN-Sequential&  0.878  & 0.874 & 0.857    \\ \hline
           Method   & jogging&running&walking  \\ \hline
         TKN&0.762 & 0.759  &  0.806      \\
         TKN-Sequential&   0.783  & 0.775 &  0.820  \\ \hline
    \end{tabular}
}
     \caption{SSIM performances on different KTH's actions.}
\label{different actions}
\end{table}
\begin{figure} [t]
	\centering
	
	\includegraphics[width=13cm]{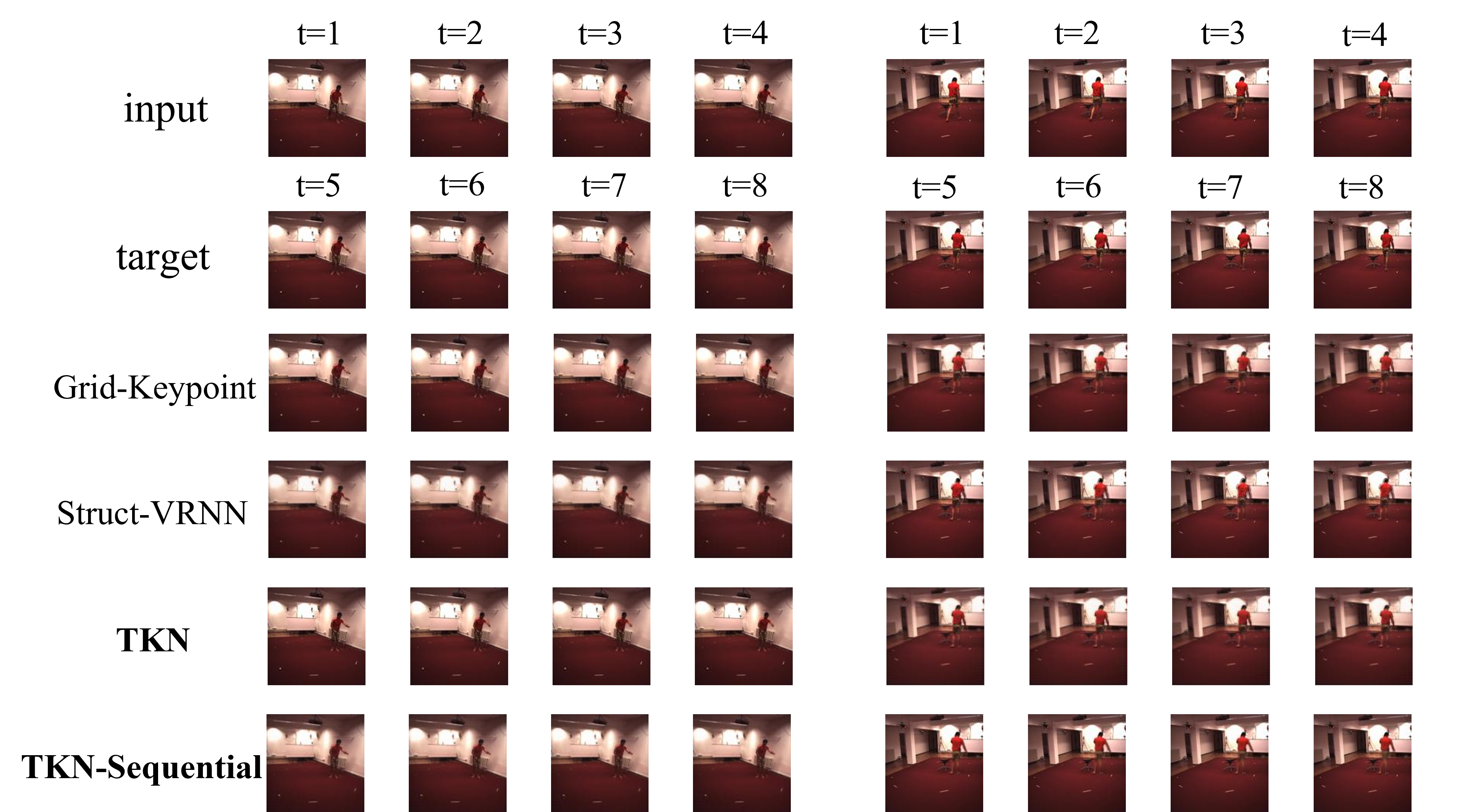}
	\caption{ Results of short-range predictions on Human3.6.}
	\label{fig:human36} 
\end{figure}

\begin{comment}
\begin{figure*} [t]
	\centering
	\subfloat[\label{fig:human1}]{
		\includegraphics[width=8cm]{figures/human_s1.png}}
    \subfloat[\label{fig:human2}]{
		\includegraphics[width=8cm]{figures/human_s2.png} }
	\caption{ Results of short-range predictions on Human3.6.}
	\label{fig:human36} 
\end{figure*}
\end{comment}
\begin{comment}
\begin{figure*}
	\begin{minipage}[t]{0.5\linewidth}
		\centering
		\includegraphics[width=8.5cm]{figures/human_s1.png}
	\end{minipage}
	\begin{minipage}[t]{0.5\linewidth}
           
		\centering
		\includegraphics[width=8.5cm]{figures/human_s2.png}
  
	\end{minipage}
	\caption{Results of short-range predictions on Human3.6.}
	\label{fig:end}
\end{figure*}
\end{comment}

\subsection{Speed and Accuracy}

\begin{table}[t]
\centering
\begin{tabular}{c|c|c|c}  \hline
         Method& MSE $\downarrow$  & MAE$\downarrow$ & SSIM$\uparrow$ \\    \hline
     ConvLSTM &  103.3  & 182.9 & 0.707   \\ 
        PredRNN &  56.8 & 126.1&   0.867      \\  
       PredRNNv2&  48.4  & -  & 0.891\\
       PhyDNet&    24.4  &  70.3  & 0.947           \\ 
       MIM &  	44.2	&101.1	&0.910 \\
       E3D-LSTM&   41.3 & 86.4  & 0.910         \\ 
       CrevNet + ConvLSTM &38.5 &  - & 0.928 \\
       CrevNet + ST-LSTM & 22.3&  - &\textbf{0.949} \\
        SimVP  &  23.8&  - &0.948  \\
      \textbf{ TKN } &   24.1  & 70.1  & \textbf{0.949}      \\   
      \textbf{TKN-Sequential} &  24.7 &  72.9 &  0.945 \\ \hline
   %\caption{The results of FLOPs and the number of parameters}
\end{tabular}
\caption{Quantitative Results on the Moving Mnist Dataset.}
\label{tab:MMnist}
\end{table}

\begin{table}[t]
\centering
\begin{tabular}{c|c|c}  \hline
         Method& SSIM$\uparrow$ & PSNR $\uparrow$   \\    \hline
     BeyondMSE &  0.847  & -   \\ 
        MCnet &  0.879 & -     \\  
       PredNet &  0.905 & 27.6 \\
       ContextVP &    0.921 & 28.7           \\ 
       rCycleGan &  	0.919 & 29.21	 \\
       CrevNet &  0.925 &29.3           \\ 
       STMFANet &  0.927 & 29.1 \\ 
       SimVP & \textbf{0.940} &33.1 \\
      \textbf{ TKN } &   0.925  &  29.6       \\   
      \textbf{TKN-Sequential} &   0.927 & 29.7   \\   \hline
   %\caption{The results of FLOPs and the number of parameters}
\end{tabular}
\caption{Quantitative Results on the Caltech Pedestrian Dataset.}
\label{tab:Caltech}
\end{table}

\begin{table*}[t!]
    \centering
     \begin{tabular}{c|c|c|c|c|c|c}   \hline
        \multirow{2}{*}{Method}  &  \multicolumn{3}{c|}{Keypoint detector } & \multicolumn{3}{c}{Predictor}   \\ 
        \cline{2-7}
        &Time (ms)  &  FLOPs (G)  & Params (M) & Time (ms)&FLOPs (G)& Params (M)  \\ \hline
  Struct-VRNN& 104&11.8 &  0.8 & 38&  \textbf{0.1} & 3.5  \\  
  Grid-Keypoint& 142&17.7  & 1.8 & 84& 8.5  & \textbf{1.7} \\     \hline
   \textbf{TKN (w/o tp) }  & 67&  \textbf{1.4} & \textbf{0.1} &  \textbf{8.3}& 0.2 &  18.9         \\  
       \textbf{TKN}   & \textbf{8.2}&  \textbf{1.4} & \textbf{0.1} &  \textbf{8.3} & 0.2 &  18.9         \\   \hline
         
\end{tabular}
    \caption{Time, FLOPs, and the number of parameters comparisons between the Keypoint-based models.}
\label{FLOPs:2}
\end{table*}

Table~\ref{tab:1} summarizes the performance comparison on KTH and Human3.6 datasets. For KTH, we input 10 frames to predict 10 frames during training and 20 frames during testing. For Human3.6, we input 4 frames to predict 4 frames during both training and testing. \emph{TIME (train)} refers to the average period length per training epoch in seconds. \emph{Time (test)} indicates the period length from inputting the frames to after generating the predicted frames in milliseconds. \emph{FPS} is the number of generated frames per second calculated via \emph{Time (test)}. \emph{Memory} indicates the maximum memory consumption at a stable status. Note that to ensure fair comparisons with the end-to-end training methods, \emph{TIME (train)} and \emph{Memory (train)} of TKN, Struct-VRNN and Grid-Keypoint in Table~\ref{tab:1} are all tested without freezing parameters. The results show that \sysname\ outperforms most baselines in both speed and memory consumption significantly with only minor accuracy deterioration on both datasets.

\vskip 0.1in \noindent\textbf{KTH} results show that \sysname\ performed 19 times faster than the best method E3D-LSTM with only 0.9\% and 5.5\% degradation in SSIM and PSNR during testing, while reducing memory consumption by at least 12.7\% (training) and 0.9\% (testing) compared to the second best methods Struct-VRNN and PredRNN. As such, \sysname\ can bear up to as large a batch size as 150 with up to 24 GB memory which no baseline can even come close to. \sysname\ is 4 times faster than TKN (w/o tp). Fig.~\ref{fig:kthlong} shows the performance of long-range prediction performance tested on the walking class of KTH, with 10 frames as input for predicting 40 frames. The result shows that the \sysname\ predicts the position and pose of a person fairly well while TKN-Sequential presents more and clearer details, because TKN only uses the background information of a fixed frame to synthesize the following frames. We  compare the performance of our models on the different action classes contained in KTH, each class with 100 randomly selected video sequences of each KTH's action class for tests. As summarized in Table \ref{different actions}, TKN-Sequential performs better than TKN on actions with large movements such as walking, jogging, and running, while TKN performs better on handwaving, handclapping, and boxing which have smaller movements.
\begin{comment}
\begin{figure}[t!]  
\centering
\includegraphics[width=.8\columnwidth]{figures/human36.png} 
\caption{ Results of short-range predictions on Human3.6. Due to space limitations, we show more comparison figure in supplements.}       
\label{fig:human}   
\end{figure}
\end{comment}

\vskip 0.1in \noindent\textbf{Human3.6} results show that \sysname\ outperforms the baselines on accuracy performance. Moreover, \sysname\ reduces time and memory consumption by 6\% and 49\% during training, and 66\% and 9\% during testing, % \pz{the decrease calculation should use the deceased value divided by the original value (the larger one)}, 
compared to the second best alternative. Figure \ref{fig:human36} depict the comparison of TKN and baselines on the Human3.6 dataset for short-range prediction, as most baselines do. The changes of the actions are small within a short period. %seem obvious due to small number of frames \pz{reviewer may ask why not put more frames for better observation of action change} \textcolor{red}{Because most baselines on Human 3.6 are testing short-range predictions}. 
But upon closer observation, we can tell that the lighting of the background and the movement of the person in TKN are closest to the goundtruth.%Figure~\ref{fig:human} verifies \sysname\ performs well on short-range predictions on Human3.6.

\begin{figure} [t]
	\centering
	
	\includegraphics[width=13cm]{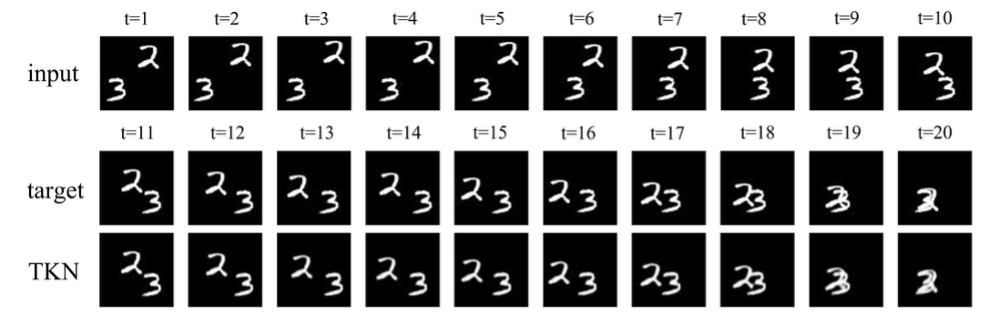}
	\caption{ Qualitative results on the Moving Mnist dataset. (10→10).}
	\label{fig:mnist} 
\end{figure}

\vskip 0.1in \noindent\textbf{Moving Mnist and Caltech Pedestrian. } We did not test the prediction speed on these two datasets since it's only related to the image shape. As shown in Fig.\ref{fig:mnist} and Fig.\ref{caltech}, TKN predicted the outcomes with good accuracy on both Moving Mnist and Caltech Pedestrian datasets. As summarized in Tab.\ref{tab:MMnist} and Tab.\ref{tab:Caltech}, TKN achieved comparable performances to the state-of-the-art (SOTA).

\begin{figure} [t]
	\centering
	
	\includegraphics[width=13cm]{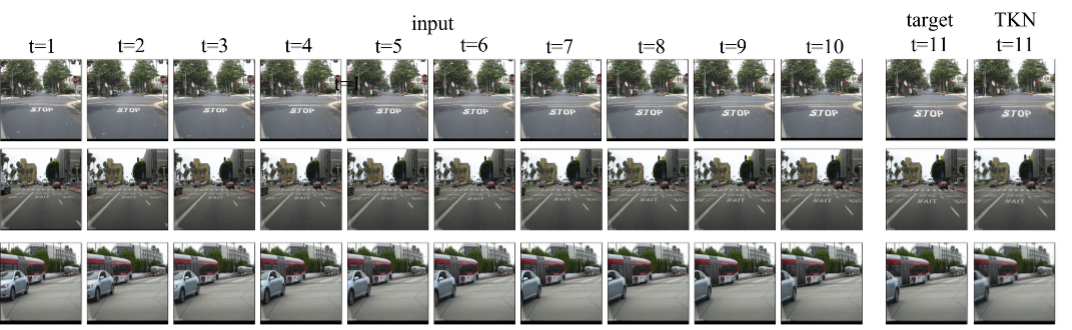}
	\caption{ Qualitative results on the Caltech Pedestrian dataset. (10→1).}
	\label{fig:caltech} 
\end{figure}

\begin{table}[t]
\centering
\begin{tabular}{c|c|c}  \hline
         Method& FLOPs (G) $\downarrow$  & Params (M)$\downarrow$   \\    \hline
     ConvLSTM &  93.7  & 4.8   \\ 
        PredRNN &   29.4 & 6.0          \\  
       PredRNNv2&  29.6  &   6.1  \\
       PhyDNet& 21.7  &\textbf{3.0}           \\
        SLAMP&  95.0    & 49.4      \\  
       E3D-LSTM&   270.2 &3.7           \\ \hline
     GridNet& 26.2 &  3.3         \\ 
      Struct-VRNN&   11.8 &4.3       \\ \hline
      \textbf{ TKN } &    \textbf{1.6}  & 19.1       \\   
      \textbf{TKN (w/o tp)} & \textbf{1.6}  &19.1   \\
      \textbf{TKN-Sequential} &   3.5 & 105  \\   \hline
   %\caption{The results of FLOPs and the number of parameters}
\end{tabular}
\caption{The results of FLOPs and the number of parameters.$\downarrow$ means the less the better.}
\label{FLOPs:1}
\end{table}

\begin{figure}[t]  
\centering
\includegraphics[width=7cm]{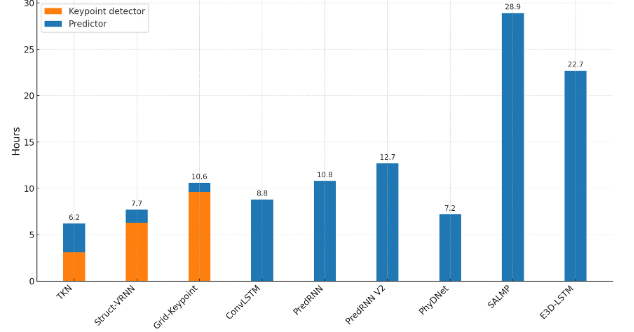}
\caption{Total training time of \sysname and baselines.}
\label{hours}  
\end{figure}

\begin{table}[t]
\centering
\setlength{\tabcolsep}{5mm}{
\begin{tabular}{c|c|c|c|c|c}  \hline
             Conv Kernel & SSIM   & PSNR  & Speed(ms)&  FLOPs(G)  & Params(M)     \\    \hline
       $3 \times 3$ &  \textbf{0.916}  & \textbf{31.91}  &  8.2  & 1.4  & 0.14    \\ 
        $2 \times 2$ &  0.862  & 28.55  &  7.6 &  0.5 &  0.06     \\ 
        $1 \times 1$ &  0.851  & 28.07  &  \textbf{7.1}   & \textbf{0.2}  &  \textbf{0.02}   \\ 
$3 \times 3$, $1 \times 1$ &  0.900  & 30.84  &  7.8 &  0.9  &    0.08     \\   \hline
   %\caption{The results of FLOPs and the number of parameters}
\end{tabular}
}
\caption{The influence of convolution kernel size on the reconstruction accuracy and inference speed of keypoint detector.  ``$3 \times 3$, $1 \times 1$'' indicates using $1 \times 1$ convolution kernel for layers that change only the channel size and not the heatmap size and $3 \times 3$ convolution kernel for the other layers.}
\label{tab:diff_conv_size}
\end{table}

\begin{table}[t!]
\small
\centering
\setlength{\tabcolsep}{6.5mm}{
    \begin{tabular}{c|c|c|c|c|c}   \hline
        \multirow{2}{*}{Method}  &  \multirow{2}{*}{Num}& \multicolumn{2}{c|}{Reconstruction} & \multicolumn{2}{c}{Prediction}   \\ 
        \cline{3-6}
       &  &   SSIM  & PSNR  & SSIM  & PSNR   \\ \hline
  Struct-VRNN& 12&0.821* &   27.86* &  0.766* &24.29*  \\  
  Grid-Keypoint& 12&0.862* &   29.68*&  0.837* &27.11*  \\    
  Separation-Net&16 &0.900 &  30.95 & 0.855 &  26.81 \\\hline
        \multirow{7}{*}{\textbf{TKN}}&4     &  0.895 & 30.43&   0.850   &  26.66           \\   
        & 8    & 0.909 &  31.28&  0.854   & 26.72               \\   
         &12    &  0.914   & 31.71   &   0.863   &  27.40           \\
         &16    & 0.916  &  31.91  & \textbf{0.871}  &   \textbf{27.71}          \\
        & 20   &    0.915  & 31.95   & 0.861   & 27.23  \\
         &32    & 0.922&  32.48 &0.858   &     26.92     \\
        &64    &\textbf{0.925}   & \textbf{32.71}  & 0.849     &     26.46          \\   \hline
         
     \end{tabular}
}
      \caption{The results of frame reconstruction and prediction using different numbers of keypoints. ``Separation-Net'' refers to the structure in Figure~\ref{fig:encoder2}. 12, 12, 16, are the number of keypoints when Struct-VRNN, Grid-Keypoint, and Separation-Net achieved their best performances, respectively.}%\pz{where's the major caption? For example, the xxx performance of xxx} }
          \label{tab:diff_num_keypoint}
\end{table}

\begin{table*}[t!]
\small
    %\vspace{-0.15in}
    \centering
   \setlength{\tabcolsep}{0.5mm}{
    \begin{tabular}{c|c|c|c|c|c|c|c|c|c}  \hline
              Method&SSIM&PSNR& TIME (s)& TIME (ms)& FPS&Memory (MB) & Memory (MB) & FLOPs(G)&Params(M)\\ 
               &      &      & train& test& test& train &test & &   \\         \hline
         TKN&  \textbf{0.871}&\textbf{27.71}& \textbf{13} &   17 & 1,176&  4,945 &\textbf{1,705}  & 1.6& 19.1        \\
         TKN+RNN& 0.826&25.61 &\textbf{13}&   15&  1,333 & 5,479 & 2,131     &1.9 &47.5                 \\
         TKN+LSTM& 0.815& 25.19& 19&   23 & 870 & 9,343 & 4,287         & 3.3&189.1        \\
         TKN+GRU& 0.829 &25.82  & 17 &   22 & 1,000 & 8,029 &   3,567   & 2.8&141.9               \\ 
        TKN+MLP& 0.814 &25.12  & \textbf{13} &   \textbf{13} & \textbf{1,538} & \textbf{4,841} &    1,733  &1.5 & 14.4               \\ \hline
    \end{tabular}
    }
    \caption{Performance comparison between TKN employing different predictors.}
    
\label{tab:diff_predictor}
\end{table*}

\begin{figure}[t!]  
\centering
\includegraphics[width=8.2cm]{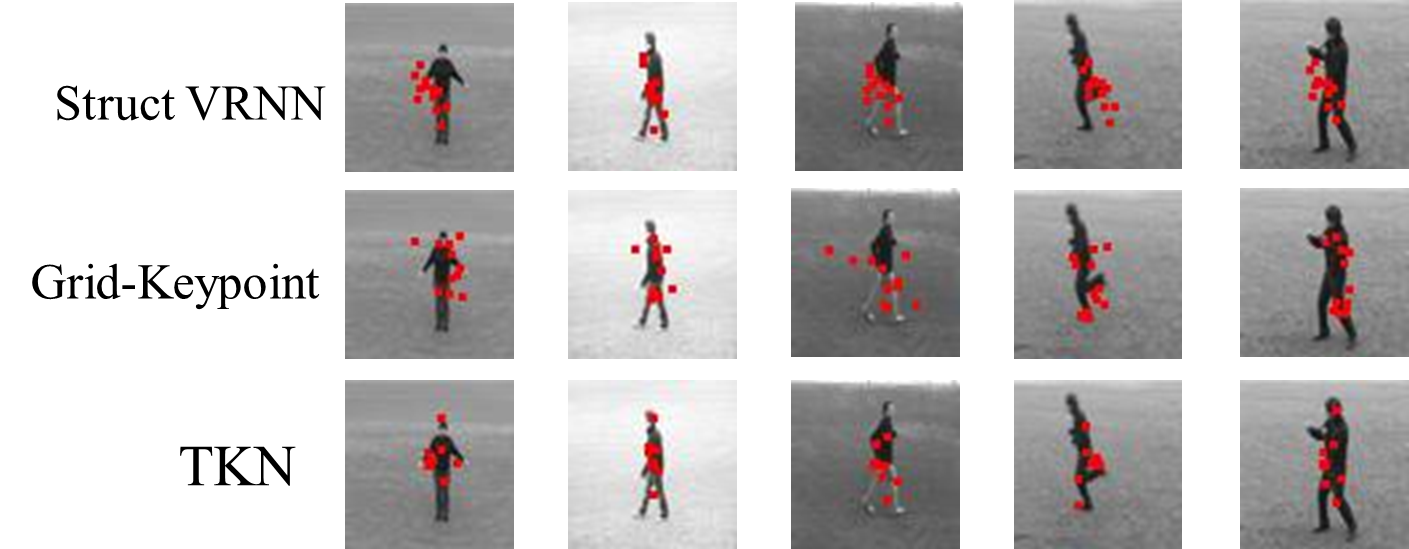} 
\caption{Comparison of the keypoints extracted by different methods}   
\label{fig:keypoints}   
\end{figure}

\begin{table*}[t!]
\centering
\resizebox{\textwidth}{8mm}{
 \begin{tabular}{c|c|c|c|c|c|c}  \hline
        Method& SSIM  & PSNR  & TIME (ms)& FPS  &Memory (MB) &TIMES (ms) \\
             &                  &                & all model(test)              & (test)         &(test)  &Predictor(test)    \\         \hline
       TKN (employs only the encoder)   &\textbf{0.871} & \textbf{27.71}    & \textbf{17}  &  \textbf{1,176} & \textbf{1,705}  &   \textbf{8.3}        \\   
     TKN (employs whole transformer) &  0.800&  25.87 &  93 &   215 & 1,759  &     74       \\\hline
\end{tabular}
} 
\caption{Performance comparison of TKN's prediction module between using only the transformer encoder and whole transformer.}
\label{tab:transformer}
\end{table*}

\vskip 0.1in \noindent\textbf{Deeper speed analysis.} To help understand why TKN runs so much faster than SOTA algorithms, we measure FLOPs and the number of parameters for each method. \sysname\ and TKN(w/o tp) share the same structure and thus have the same numbers of FLOPs and parameters. As shown in Table \ref{FLOPs:1}, \sysname\  and TKN-Sequential have much fewer FLOPs than the baselines, indicating their much higher computation efficiencies. Table~\ref{FLOPs:2} summarizes the detailed comparison between TKN, TKN(w/o tp), and the other two Keypoint-based methods. For the choice of predictor, \sysname\ uses transformer encoder while Grid-Keypoint uses convlstm and Struct-VRNN uses VRNN. The results in Table~\ref{FLOPs:2} show that \sysname\ presents the fastest speed in both modules and provides an 8 times speedup compared with TKN(w/o tp), indicating the key role of keypoint detector in terms of prediction speed and the advantage of parallel scheme. We can also find that although the predictor of TKN has more FLOPs and number of parameters due to the larger number of parameters of the employed transformer encoder~\cite{vaswani2017attention}, it runs about 5 to 10 times faster than the others. 
%That is because the transformer encoder can be calculated in parallel while Struct-VRNN uses the RNN's step-by-step loop method.% As for the number of parameters, because the transformer encoder and the transformer structure have large numbers of parameters \cite{vaswani2017attention} as shown in Table \ref{FLOPs:2},\sysname\ and TKN-Long do not have as few parameters as other baselines.

\vskip 0.1in \noindent\textbf{Overall training time.} We compare the overall training time considering various numbers of required epochs before convergence. Note that here \sysname, Grid-Keypoint, and Struct-VRNN use the two-step training as mentioned in Section~\ref{sec:setup}. Although TKN's predictor trains slower than Grid-Keypoint and Struct-VRNN because its Transformer encoder takes 750 epochs to reach the optima while Convlstm and VRNN takes only 20 and 50, TKN's overall training speed is up to 2 to 3 times faster than the baselines as shown in Figure \ref{hours}.

\subsection{Ablation Experiments}
\vskip 0.1in \noindent\textbf{Keypoint Detector.} We test the impact of convolution kernel size and find that it presents a much larger impact on reconstruction accuracy than on inference speed of keypoint detector as summarized in Table~\ref{tab:diff_conv_size}. Therefore we choose to use the $3 \times 3$ convolution kernel.

% We then compare the influence of different structures and numbers of keypoints on reconstruction and prediction. 
Table~\ref{tab:diff_num_keypoint} shows that \sysname\ reconstructs frames better than the two Keypoints-based baselines and the sequential structure in Fig.~\ref{fig:encoder2}, and achieves better performance with more keypoints. Meanwhile, the prediction module hits a performance bottleneck at 20 keypoints, indicating that too many keypoints pose difficulties for prediction. Figure~\ref{fig:keypoints} shows that the keypoint detector of \sysname\ performs better at capturing dynamic information than that of Struct-VRNN and Grid-keypoint on different actions.% \pz{worse than Grid-keypoint?} \textcolor{red}{forgot to write}.

\vskip 0.1in \noindent\textbf{Predictor} relies on the results of the keypoint detector. We verified its performance by replacing its Transformer encoder with alternative modules, i.e., RNN~\cite{elman1990finding}, LSTM~\cite{hochreiter1997long},  GRU~\cite{cho2014learning}, and MLP, which are widely used in prediction tasks. We adjusted the input dimension to $\mathbb{R}^{d_{model}} $ (512), number of layers to 6, and the dimension of the hidden layers to 2048, in all modules, for fair comparisons. We use the encapsulated RNN, LSTM, and GRU modules from PyTorch and  use the MLP structure in Mlp-mixer\cite{tolstikhin2021mlp}). As shown in Table~\ref{tab:diff_predictor}, our predictor module presents comparable training and testing speeds with significantly higher accuracy and memory efficiency.

\begin{table}[t]
\centering
\setlength{\tabcolsep}{5mm}{
\begin{tabular}{c|c|c|c|c|c}  \hline
          Method    & SSIM   & PSNR  & Speed (ms) &  FLOPs(G)  & Params(M)\\    \hline
      explicit     &  0.852  & 26.50  &  \textbf{6.7} &     \textbf{0.01}   & \textbf{0.7}     \\ 
      latent   &  \textbf{0.871}  & \textbf{27.71} &  8.3  &   0.19     & 18.9      \\   \hline
   %\caption{The results of FLOPs and the number of parameters}
\end{tabular}
}
\caption{Comparison between using explicit and latent representation of keypoints.}
\label{tab:implicit}
\end{table}
%这实际上是必然的，因为transformer最开始是解决自然语言处理方向的问题，在使用它之前需要对所有的词标注id,然后对id进行嵌入处理，嵌入向量和词语id是一一对应的，它的翻译流程实际上与RNN的循环原理是相似的，每次输出的高维结果会与嵌入向量对比然后得到相应的词语id（一般是取与输出相似度最高的嵌入向量），然后再将这个词语id的嵌入向量作为输入，送到transformer中去翻译下一个词。也就是说他们的任务输入和输出都是有限的离散量，而我们的任务预测的keypoints每个元素实际上都是连续的量，也就是说我们没办法用有限的id来标注，所以在翻译过程用中就没有从高维变量到id那一个步骤，对于transformer如果翻译正确找到正确的id，那么这一次翻译的正确率就是100%,下次的输入用的是正确的id的嵌入向量，而我们的任务每次翻译不可能100%正确的输出浮点指，所以在下一次翻译的时候，输入就会有误差，这样子整个过程下来就会造成错误的累加
\par Table~\ref{tab:transformer} compares the performance of TKN's predictor employing the complete transformer structure with when using only its encoder part. We can see that the encoder-only method works much better in terms of both prediction speed and accuracy. This is because the transformer initially proposed for NLP problems requires embedding each word's label ID into a vector. Its translation process is similar to RNN's cycle principle which compares each high-dimensional output with the embedding vectors to get the word ID, and then inputs the corresponding embedding vector to the transformer to translate the next word. In short, their input and output are finite discrete quantities, while the keypoints in our prediction task are continuous quantities which cannot be labeled with finite IDs, hence excluding the possibility of mapping the high-dimensional output to IDs. Moreover, each output in our task, which is the next input, is a floating point which cannot be acquired with 100\% accuracy. Thus, the small errors in each transformer cycle are accumulated.
%And if the transformer translates well  and it can find the correct id, then the next input is the embedding vector with the correct id, This id and its embedding vector as input are 100\% correct and our task cannot output the floating point finger with 100\% correct translation each time, so the input will have error in the next translation, which will cause the accumulation of errors in the whole process.
%\pz{edition currently stop here}
TKN employing the complete transformer has a long prediction time because the translation part of the transformer is a step-by-step process and each step goes through a complete transformer, while encoder-only TKN outputs all the results in one step.

Further, we test the impact of using explicit or latent representation of keypoints. As shown in Table~\ref{tab:implicit}, latent representation in high-dimension presents higher prediction accuracy.
On the other hand, explicit representation only has limited speed improvement albeit it has fewer FLOPs and parameters.
\par %We also provide a comparison video of TKN and baselines in the attached file. The first second shows the observed content, and the later seconds show the predicted content. It can be seen that TKN and TKN-Long have clearer and more accurate video content than baselines. And TKN-Long is more accurate in details and shows more precise position predictions than TKN.%

\section{Conclusion}
This paper has presented \sysname, a video prediction model that combines the advantages of keypoints and transformer models to achieve comparable accuracy performance with SOTA but with significantly less time and memory cost. TKN realizes real-time video prediction for the first time and opens the door for numerous futuristic applications demanding such capabilities. 
For future work, we plan to combine TKN with new AR applications and apply it to multi-person videos with higher resolutions.

\bibliographystyle{ACM-Reference-Format}
\bibliography{main}

@String{Computing = "Computing" }

@String{Computer = "{IEEE} Computer" }

@String{Springer = "Springer-Verlag" }

@article{meng2018zero,
  title={Zero-shot learning via robust latent representation and manifold regularization},
  author={Meng, Min and Yu, Jun},
  journal={IEEE Transactions on Image Processing},
  volume={28},
  number={4},
  pages={1824--1836},
  year={2018},
  publisher={IEEE}
}

@article{chen2020long,
  title={Long-term video prediction via criticization and retrospection},
  author={Chen, Xinyuan and Xu, Chang and Yang, Xiaokang and Tao, Dacheng},
  journal={IEEE Transactions on Image Processing},
  volume={29},
  pages={7090--7103},
  year={2020},
  publisher={IEEE}
}

@article{tao2020latent,
  title={Latent complete row space recovery for multi-view subspace clustering},
  author={Tao, Hong and Hou, Chenping and Qian, Yuhua and Zhu, Jubo and Yi, Dongyun},
  journal={IEEE Transactions on Image Processing},
  volume={29},
  pages={8083--8096},
  year={2020},
  publisher={IEEE}
}

@article{zhou2021latent,
  title={Latent correlation representation learning for brain tumor segmentation with missing MRI modalities},
  author={Zhou, Tongxue and Canu, St{\'e}phane and Vera, Pierre and Ruan, Su},
  journal={IEEE Transactions on Image Processing},
  volume={30},
  pages={4263--4274},
  year={2021},
  publisher={IEEE}
}

@inproceedings{mcgehee2000driver,
  title={Driver reaction time in crash avoidance research: Validation of a driving simulator study on a test track},
  author={McGehee, Daniel V and Mazzae, Elizabeth N and Baldwin, GH Scott},
  booktitle={Proceedings of the human factors and ergonomics society annual meeting},
  volume={44},
  number={20},
  pages={3--320},
  year={2000},
  organization={Sage Publications Sage CA: Los Angeles, CA}
}

@inproceedings{wang2018eidetic,
  title={Eidetic 3d lstm: A model for video prediction and beyond},
  author={Wang, Yunbo and Jiang, Lu and Yang, Ming-Hsuan and Li, Li-Jia and Long, Mingsheng and Fei-Fei, Li},
  booktitle={International conference on learning representations},
  year={2018}
}

@inproceedings{guen2020disentangling,
  title={Disentangling physical dynamics from unknown factors for unsupervised video prediction},
  author={Guen, Vincent Le and Thome, Nicolas},
  booktitle={Proceedings of the IEEE/CVF Conference on Computer Vision and Pattern Recognition},
  pages={11474--11484},
  year={2020}
}

@article{minderer2019unsupervised,
  title={Unsupervised learning of object structure and dynamics from videos},
  author={Minderer, Matthias and Sun, Chen and Villegas, Ruben and Cole, Forrester and Murphy, Kevin P and Lee, Honglak},
  journal={Advances in Neural Information Processing Systems},
  volume={32},
  year={2019}
}

@article{jakab2018conditional,
  title={Conditional image generation for learning the structure of visual objects},
  author={Jakab, Tomas and Gupta, Ankush and Bilen, Hakan and Vedaldi, Andrea},
  journal={methods},
  volume={43},
  pages={44},
  year={2018}
}

@article{wang2021predrnn,
  title={PredRNN: A recurrent neural network for spatiotemporal predictive learning},
  author={Wang, Yunbo and Wu, Haixu and Zhang, Jianjin and Gao, Zhifeng and Wang, Jianmin and Yu, Philip S and Long, Mingsheng},
  journal={arXiv preprint arXiv:2103.09504},
  year={2021}
}

@article{wang2017predrnn,
  title={Predrnn: Recurrent neural networks for predictive learning using spatiotemporal lstms},
  author={Wang, Yunbo and Long, Mingsheng and Wang, Jianmin and Gao, Zhifeng and Yu, Philip S},
  journal={Advances in neural information processing systems},
  volume={30},
  year={2017}
}

@inproceedings{ying2018better,
  title={Better guider predicts future better: Difference guided generative adversarial networks},
  author={Ying, Guohao and Zou, Yingtian and Wan, Lin and Hu, Yiming and Feng, Jiashi},
  booktitle={Asian Conference on Computer Vision},
  pages={277--292},
  year={2018},
  organization={Springer}
}

@article{shi2015convolutional,
  title={Convolutional LSTM network: A machine learning approach for precipitation nowcasting},
  author={Shi, Xingjian and Chen, Zhourong and Wang, Hao and Yeung, Dit-Yan and Wong, Wai-Kin and Woo, Wang-chun},
  journal={Advances in neural information processing systems},
  volume={28},
  year={2015}
}

@article{vaswani2017attention,
  title={Attention is all you need},
  author={Vaswani, Ashish and Shazeer, Noam and Parmar, Niki and Uszkoreit, Jakob and Jones, Llion and Gomez, Aidan N and Kaiser, {\L}ukasz and Polosukhin, Illia},
  journal={Advances in neural information processing systems},
  volume={30},
  year={2017}
}

@inproceedings{liu2022video,
  title={Video swin transformer},
  author={Liu, Ze and Ning, Jia and Cao, Yue and Wei, Yixuan and Zhang, Zheng and Lin, Stephen and Hu, Han},
  booktitle={Proceedings of the IEEE/CVF Conference on Computer Vision and Pattern Recognition},
  pages={3202--3211},
  year={2022}
}

@inproceedings{liu2021swin,
  title={Swin transformer: Hierarchical vision transformer using shifted windows},
  author={Liu, Ze and Lin, Yutong and Cao, Yue and Hu, Han and Wei, Yixuan and Zhang, Zheng and Lin, Stephen and Guo, Baining},
  booktitle={Proceedings of the IEEE/CVF International Conference on Computer Vision},
  pages={10012--10022},
  year={2021}
}

@inproceedings{liang2021swinir,
  title={Swinir: Image restoration using swin transformer},
  author={Liang, Jingyun and Cao, Jiezhang and Sun, Guolei and Zhang, Kai and Van Gool, Luc and Timofte, Radu},
  booktitle={Proceedings of the IEEE/CVF International Conference on Computer Vision},
  pages={1833--1844},
  year={2021}
}

@article{dosovitskiy2020image,
  title={An image is worth 16x16 words: Transformers for image recognition at scale},
  author={Dosovitskiy, Alexey and Beyer, Lucas and Kolesnikov, Alexander and Weissenborn, Dirk and Zhai, Xiaohua and Unterthiner, Thomas and Dehghani, Mostafa and Minderer, Matthias and Heigold, Georg and Gelly, Sylvain and others},
  journal={arXiv preprint arXiv:2010.11929},
  year={2020}
}

@inproceedings{arnab2021vivit,
  title={Vivit: A video vision transformer},
  author={Arnab, Anurag and Dehghani, Mostafa and Heigold, Georg and Sun, Chen and Lu{\v{c}}i{\'c}, Mario and Schmid, Cordelia},
  booktitle={Proceedings of the IEEE/CVF International Conference on Computer Vision},
  pages={6836--6846},
  year={2021}
}

@inproceedings{gao2021accurate,
  title={Accurate Grid Keypoint Learning for Efficient Video Prediction},
  author={Gao, Xiaojie and Jin, Yueming and Dou, Qi and Fu, Chi-Wing and Heng, Pheng-Ann},
  booktitle={2021 IEEE/RSJ International Conference on Intelligent Robots and Systems (IROS)},
  pages={5908--5915},
  year={2021},
  organization={IEEE}
}

@article{liu2021video,
  title={Video swin transformer},
  author={Liu, Ze and Ning, Jia and Cao, Yue and Wei, Yixuan and Zhang, Zheng and Lin, Stephen and Hu, Han},
  journal={arXiv preprint arXiv:2106.13230},
  year={2021}
}

@inproceedings{ronneberger2015u,
  title={U-net: Convolutional networks for biomedical image segmentation},
  author={Ronneberger, Olaf and Fischer, Philipp and Brox, Thomas},
  booktitle={International Conference on Medical image computing and computer-assisted intervention},
  pages={234--241},
  year={2015},
  organization={Springer}
}

@article{rasmus2015semi,
  title={Semi-supervised learning with ladder networks},
  author={Rasmus, Antti and Berglund, Mathias and Honkala, Mikko and Valpola, Harri and Raiko, Tapani},
  journal={Advances in neural information processing systems},
  volume={28},
  year={2015}
}

@inproceedings{oliu2018folded,
  title={Folded recurrent neural networks for future video prediction},
  author={Oliu, Marc and Selva, Javier and Escalera, Sergio},
  booktitle={Proceedings of the European Conference on Computer Vision (ECCV)},
  pages={716--731},
  year={2018}
}

@article{h36m_pami,
author = {Ionescu, Catalin and Papava, Dragos and Olaru, Vlad and Sminchisescu,  Cristian},
title = {Human3.6M: Large Scale Datasets and Predictive Methods for 3D Human Sensing in Natural Environments},
journal = {IEEE Transactions on Pattern Analysis and Machine Intelligence},
publisher = {IEEE Computer Society},
volume = {36},
number = {7},
pages = {1325-1339},
month = {jul},
year = {2014}
}

@inproceedings{akan2021slamp,
  title={SLAMP: Stochastic Latent Appearance and Motion Prediction},
  author={Akan, Adil Kaan and Erdem, Erkut and Erdem, Aykut and G{\"u}ney, Fatma},
  booktitle={Proceedings of the IEEE/CVF International Conference on Computer Vision},
  pages={14728--14737},
  year={2021}
}

@article{cho2014learning,
  title={Learning phrase representations using RNN encoder-decoder for statistical machine translation},
  author={Cho, Kyunghyun and Van Merri{\"e}nboer, Bart and Gulcehre, Caglar and Bahdanau, Dzmitry and Bougares, Fethi and Schwenk, Holger and Bengio, Yoshua},
  journal={arXiv preprint arXiv:1406.1078},
  year={2014}
}

@article{elman1990finding,
  title={Finding structure in time},
  author={Elman, Jeffrey L},
  journal={Cognitive science},
  volume={14},
  number={2},
  pages={179--211},
  year={1990},
  publisher={Wiley Online Library}
}

@article{hochreiter1997long,
  title={Long short-term memory},
  author={Hochreiter, Sepp and Schmidhuber, J{\"u}rgen},
  journal={Neural computation},
  volume={9},
  number={8},
  pages={1735--1780},
  year={1997},
  publisher={MIT Press}
}

@inproceedings{schuldt2004recognizing,
  title={Recognizing human actions: a local SVM approach},
  author={Schuldt, Christian and Laptev, Ivan and Caputo, Barbara},
  booktitle={Proceedings of the 17th International Conference on Pattern Recognition, 2004. ICPR 2004.},
  volume={3},
  pages={32--36},
  year={2004},
  organization={IEEE}
}

@article{denton2017unsupervised,
  title={Unsupervised learning of disentangled representations from video},
  author={Denton, Emily L and others},
  journal={Advances in neural information processing systems},
  volume={30},
  year={2017}
}

@inproceedings{castrejon2019improved,
  title={Improved conditional vrnns for video prediction},
  author={Castrejon, Lluis and Ballas, Nicolas and Courville, Aaron},
  booktitle={Proceedings of the IEEE/CVF International Conference on Computer Vision},
  pages={7608--7617},
  year={2019}
}

@article{xu2019unsupervised,
  title={Unsupervised discovery of parts, structure, and dynamics},
  author={Xu, Zhenjia and Liu, Zhijian and Sun, Chen and Murphy, Kevin and Freeman, William T and Tenenbaum, Joshua B and Wu, Jiajun},
  journal={arXiv preprint arXiv:1903.05136},
  year={2019}
}

@inproceedings{blattmann2021understanding,
  title={Understanding object dynamics for interactive image-to-video synthesis},
  author={Blattmann, Andreas and Milbich, Timo and Dorkenwald, Michael and Ommer, Bjorn},
  booktitle={Proceedings of the IEEE/CVF Conference on Computer Vision and Pattern Recognition},
  pages={5171--5181},
  year={2021}
}

@inproceedings{wang2018predrnn++,
  title={Predrnn++: Towards a resolution of the deep-in-time dilemma in spatiotemporal predictive learning},
  author={Wang, Yunbo and Gao, Zhifeng and Long, Mingsheng and Wang, Jianmin and Philip, S Yu},
  booktitle={International Conference on Machine Learning},
  pages={5123--5132},
  year={2018},
  organization={PMLR}
}

@article{wang2022predrnn,
  title={Predrnn: A recurrent neural network for spatiotemporal predictive learning},
  author={Wang, Yunbo and Wu, Haixu and Zhang, Jianjin and Gao, Zhifeng and Wang, Jianmin and Philip, S Yu and Long, Mingsheng},
  journal={IEEE Transactions on Pattern Analysis and Machine Intelligence},
  volume={45},
  number={2},
  pages={2208--2225},
  year={2022},
  publisher={IEEE}
}

@inproceedings{zhao2020rnn,
  title={Do rnn and lstm have long memory?},
  author={Zhao, Jingyu and Huang, Feiqing and Lv, Jia and Duan, Yanjie and Qin, Zhen and Li, Guodong and Tian, Guangjian},
  booktitle={International Conference on Machine Learning},
  pages={11365--11375},
  year={2020},
  organization={PMLR}
}

@article{tolstikhin2021mlp,
  title={Mlp-mixer: An all-mlp architecture for vision},
  author={Tolstikhin, Ilya O and Houlsby, Neil and Kolesnikov, Alexander and Beyer, Lucas and Zhai, Xiaohua and Unterthiner, Thomas and Yung, Jessica and Steiner, Andreas and Keysers, Daniel and Uszkoreit, Jakob and others},
  journal={Advances in Neural Information Processing Systems},
  volume={34},
  pages={24261--24272},
  year={2021}
}

%%
%% If your work has an appendix, this is the place to put it.
\appendix

\end{document}